\documentclass[dvipsnames, 11pt]{article}
\usepackage{inconsolata}
\usepackage[preprint]{acl}

\usepackage{times}
\usepackage{latexsym}
\usepackage[T1]{fontenc}
\usepackage[utf8]{inputenc}
\usepackage{microtype}
\usepackage[normalem]{ulem}
\usepackage{inconsolata}
\usepackage[smartEllipses]{markdown}
\usepackage{xfrac}
\usepackage{colortbl}
\usepackage{adjustbox}
\usepackage{diagbox}
\usepackage{amsmath,amssymb, amsthm}
\usepackage[inline]{enumitem}
\usepackage{array,multirow,graphicx,makecell}
\usepackage{booktabs}
\usepackage{tabularx}
\usepackage{xfrac}
\usepackage[skins]{tcolorbox}
\tcbuselibrary{breakable}
\usepackage{framed}
\usepackage{tabto}
\usepackage{tabulary}
\usepackage{algpseudocode}
\usepackage{algorithm}
\usepackage{listings}
\usepackage{accents}
\usepackage{anyfontsize}
\usepackage{pdflscape}
\usepackage{longtable}
\usepackage{float}
\usetikzlibrary{positioning, arrows.meta}
\usepackage{caption}
\usepackage{subcaption}

\usepackage[toc,page,header]{appendix}
\usepackage{minitoc}

\newcounter{tcbcounter}

\usepackage{amsmath,amsfonts,bm}

\def\eqref#1{equation~\ref{#1}}

\def\1{\bm{1}}

\DeclareMathAlphabet{\mathsfit}{\encodingdefault}{\sfdefault}{m}{sl}
\SetMathAlphabet{\mathsfit}{bold}{\encodingdefault}{\sfdefault}{bx}{n}

\newcommand{\E}{\mathbb{E}}

\newtheorem{definition}{Definition}

\newtcolorbox[auto counter, number within=section]{prompt}[3][]{%
  enhanced,
  breakable,
  colback=#2!5!white,
  colframe=#2!75!black,
  title=\textbf{Box \thetcbcounter: #3},
  fontupper=\footnotesize\fontfamily{cmr}\selectfont,
  #1
}

\newtcolorbox[auto counter, number within=section]{defprompt}[3][]{%
  enhanced,
  colback=#2!5!white,
  colframe=#2!75!black,
  title=\textbf{Box \thetcbcounter: #3},
  fontupper=\footnotesize\fontfamily{cmr}\selectfont,
  #1
}

\title{LIBERTy: A Causal Framework for Benchmarking Concept-Based Explanations of LLMs with Structural Counterfactuals}

\doparttoc 
\faketableofcontents 

\author{
\textbf{Gilat Toker}\textsuperscript{*},
\textbf{Nitay Calderon}\textsuperscript{*},
\textbf{Ohad Amosy},
\textbf{Roi Reichart}
\\
Faculty of Data and Decision Sciences, Technion -- Israel Institute of Technology
\\
\texttt{\{gilatt, nitay\}@campus.technion.ac.il},
\texttt{roiri@technion.ac.il}
\\[0.5ex]
\textsuperscript{*}Second author supervised the project and led the writing.
}

\begin{document}
\maketitle
\begin{abstract}
Concept-based explanations quantify how high-level concepts (e.g., gender or experience) influence model behavior, which is crucial for decision-makers in high-stakes domains. Recent work evaluates the faithfulness of such explanations by comparing them to reference causal effects estimated from counterfactuals. In practice, existing benchmarks rely on costly human-written counterfactuals that serve as an imperfect proxy. To address this, we introduce a framework for constructing datasets containing structural counterfactual pairs: LIBERTy (\textbf{L}LM-based \textbf{I}nterventional \textbf{B}enchmark for \textbf{E}xplainability with \textbf{R}eference \textbf{T}argets). LIBERTy is grounded in explicitly defined Structured Causal Models (SCMs) of the text generation, interventions on a concept propagate through the SCM until an LLM generates the counterfactual.
We introduce three datasets (disease detection, CV screening, and workplace violence prediction) together with a new evaluation metric, order-faithfulness. Using them, we evaluate a wide range of methods across five models and identify substantial headroom for improving concept-based explanations. LIBERTy also enables systematic analysis of model sensitivity to interventions: we find that proprietary LLMs show markedly reduced sensitivity to demographic concepts, likely due to post-training mitigation. Overall, LIBERTy provides a much-needed benchmark for developing faithful explainability methods.\footnote{\url{https://github.com/GilatToker/Liberty-benchmark}}
\end{abstract}

\section{Introduction}

AI systems, especially Large Language Models (LLMs), increasingly drive decisions in sensitive and high-stakes domains where \textit{textual input} plays a central role, such as finance, education, healthcare, and law \citep{RiccardoGuidotti2018,EsmaBalkir2022,llmEducation,llmLaw,SiwenLuo2024,llmFinance,benkirane2025diagnose}. 
The decisions of these opaque systems are difficult to explain, making explainability a central research challenge \citep{RiccardoGuidotti2018, EsmaBalkir2022, SiwenLuo2024}.
Among the many approaches to explainability, \textit{concept-based methods} are particularly relevant when the stakeholders are decision-makers and end-users \citep{CalderonReichart2024}. These methods focus on quantifying the influence of high-level, human-interpretable concepts, such as gender, race, or professional experience, on model predictions \citep{TCAV_Kim2018, SLEARNER, Human_KeyConcepts, feder2021, CPM, Gat2024}.

\begin{figure*}[t]
    \centering
    \includegraphics[width=0.98\textwidth]{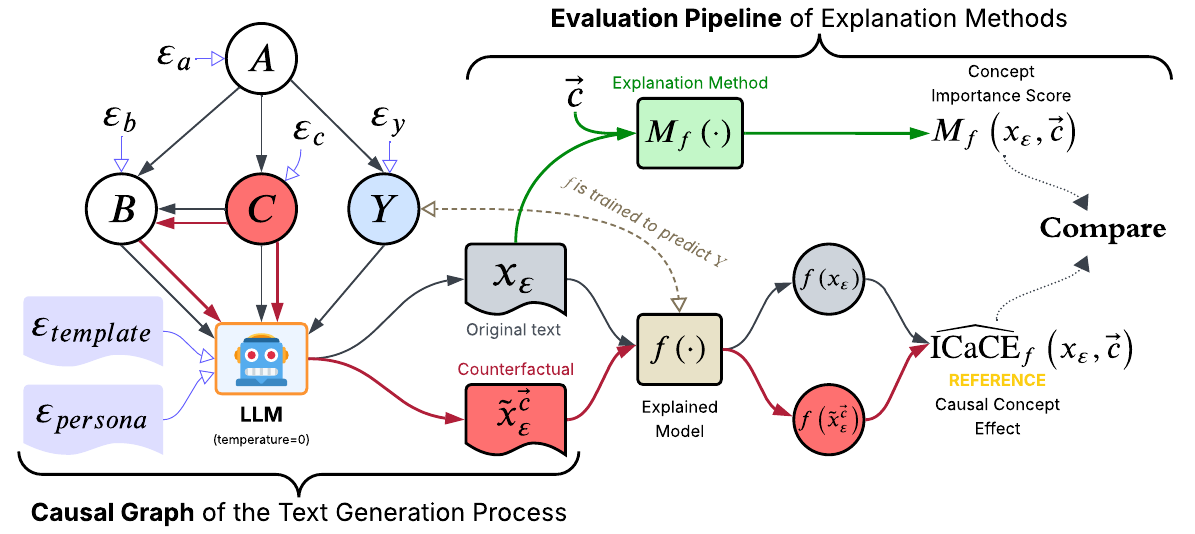}
    \caption{\textbf{Illustration of LIBERTy:} The goal is to evaluate an explanation method $M_{f}$ that explains the impact of changing a concept $C$ (by $\stackrel{\rightarrow}{c}$) on model $f$.
    \textbf{Left:} The causal graph representing the text generation process. Exogenous noise variables are denoted by $\varepsilon$, while the endogenous variables (in this illustration) are the concepts $A,B,C,Y$, the LLM-generated text $x_{\varepsilon}$, and the model prediction $f(x_{\varepsilon})$. The process of generating a structural counterfactual for the change $\stackrel{\rightarrow}{c}$ is highlighted in red: $C$ is assigned a new value and propagated through the causal graph (with $\varepsilon$ fixed) until the LLM generates the counterfactual $x^{\stackrel{\rightarrow}{c}}_{\varepsilon}$. 
    \textbf{Right:} The explanation $M_{f}(x_{\varepsilon}, \stackrel{\rightarrow}{c})$ is compared against the refrence individual causal concept effect (ICaCE), defined as the difference between $f(x^{\stackrel{\rightarrow}{c}}_{\varepsilon})$ and $f(x_\varepsilon)$.}
    \label{fig:pipeline}
    \vspace{-0.8em}
\end{figure*}

Recent studies emphasize that explanations lacking a causal basis often fail to achieve true faithfulness \citep{lyu-etal-2024-towards, Gat2024, Yeo2024}.
In causality, a causal graph encodes concepts as variables and their relationships as edges \citep{pearl2009causality}. This structure enables us to identify the roles of concepts, such as confounders, mediators, and colliders, and to estimate the causal effect of a target concept on the model \citep{cebab2022}. Despite progress at the intersection of AI and causality \citep{adjustments, feder2021, interventions, matching}, a fundamental challenge remains: evaluating whether an explanation is faithful requires comparing it to the true underlying causal mechanisms. In practice, the ground-truth mechanism is inaccessible, leaving us without a reliable benchmark for explainability methods.

One approach to address this benchmarking challenge was introduced by \citet{cebab2022}. They propose using an \textit{interventional dataset} as a systematic framework for evaluation of explanations. In their interventional dataset, CEBaB, each test example is paired with a human-written counterfactual generated by modifying a concept. The individual causal concept effect (ICaCE) is then estimated by contrasting the model's outputs on the original text and its counterfactual. Explanation methods are evaluated by comparing their importance score (of the concept) against the estimated causal effect. While CEBaB represents a significant step toward causal evaluation, it remains limited, especially given LLMs' current capabilities. First, CEBaB is confined to sentiment analysis of restaurant reviews, which are short, simple texts. Second, its causal graph comprises only four concepts, with simple relationships (no hierarchical structure). Finally, the counterfactuals are written by human annotators rather than arising from actual interventions in the \textit{data-generating process (DGP)}. Consequently, the causal effect references used as ``ground truth'' for evaluation are themselves approximations of some unobserved effects.

In this work, we address these limitations by introducing a novel framework for generating interventional datasets with structural counterfactuals that define reference causal effects: LIBERTy (\textbf{L}LM-based \textbf{I}nterventional \textbf{B}enchmark for \textbf{E}xplainability with \textbf{R}eference \textbf{T}argets).
LIBERTy, illustrated in Figure~\ref{fig:pipeline}, is based on a simple yet effective idea: explicitly defining a structured causal model (SCM) for text generation. In this framework, the LLM is a component of the SCM that instantiates concepts as natural language text. To make LLM outputs more diverse and realistic, we provide it with grounding context, such as templated real-world text and author persona, which act as exogenous noise variables in the SCM. Counterfactuals are generated by intervening on a concept (assigning it a new value) and propagating this change through the SCM until the LLM produces the corresponding counterfactual. As a result, LIBERTy provides structural counterfactuals\footnote{Formally, LIBERTy counterfactuals are gold with respect to the DGP (which the LLM that generates text is part of): given the SCM and observed exogenous values, they are generated via Pearl's three-step procedure \citep{pearl2013structural}. However, since the DGP itself and the resulting texts are synthetic, we use the term silver to refer to these structural counterfactuals. Yet, as LLMs generate an increasing share of real-world data, such setups are both common and practically meaningful.}, eliminating the need for costly human annotations and ensuring alignment between the evaluation reference target and the DGP.

LIBERTy comprises three datasets, each designed around a major societal challenge: disease detection, CV screening, and workplace violence prediction. 
We also propose a new evaluation measure, order-faithfulness, that quantifies how well an explanation method captures the relative ordering of effects induced by concept interventions. This makes it suitable for evaluating explanation methods that provide importance scores on arbitrary scales, rather than direct causal effect estimates. 

Using LIBERTy, we conduct extensive experiments to explain five NLP models and LLMs. We benchmark a range of concept-based explanation methods, including linear erasure, counterfactual generation, matching, and concept attributions. Our results show that matching methods based on representations from a dataset-specific fine-tuned model perform best overall. Still, we find substantial headroom for improvement, highlighting the need for continued explainability research.

Besides evaluating explanations, LIBERTy enables us to analyze the sensitivity of each explained model to concept interventions. For example, when the model predicts a candidate's qualification based on their CV, we examine how its prediction changes when we intervene on the candidate's race, and compare this change to the effect specified in the SCM. Our results show that fine-tuned models can track the ground-truth effects of the data. In contrast, some LLMs (like GPT-4o) exhibit very low sensitivity to demographic concepts, potentially due to dedicated post-training alignment. 

Overall, our study represents an important step toward addressing the long-standing challenge of explainability evaluation. By introducing LIBERTy, we provide researchers with a reliable, scalable, and flexible causal framework for benchmark generation, paving the way for the development of more faithful explainability methods.
\section{Related Work}

\paragraph{Concept-based Explainability}~Concept-based explainability encompasses methods that quantify the extent to which high-level, human-interpretable concepts (features, attributes, variables, rubrics) that can be explicitly or implicitly conveyed in the text influence model predictions. 
This is in contrast to token-level explanations, which emphasize tokens through techniques such as attribution or attention scores \citep{CalderonReichart2024, SiwenLuo2024, Zhao2024}. Concept-based explanations naturally align with human cognitive processes \citep{Alqaraawi2020, Kim2023, Poeta2023} and simplify the complexity inherent in lengthy textual inputs, making explanations more intuitive and easier to communicate \citep{CalderonReichart2024}. Moreover, they naturally support both local and global explanations. 
These advantages have driven their widespread use in applications such as bias detection \citep{detect_bias}, providing clear and actionable explanations \citep{good_explenation}, discovering new hidden concepts \citep{find_aspects}, explaining human preferences, reward models, and LLM-as-Judges \citep{multi_domain}, and detecting dementia \citep{dementia}. 

Among the most prominent approaches of concept-based explainability, there are \textit{Attribution methods} \citep{Ribeiro2016, Lundberg2017, TCAV_Kim2018, Human_KeyConcepts}, \textit{Concept Erasure methods} \citep{ravfogel2022linear, belrose2023leace}, \textit{Counterfactual Generation methods},~\citep{feder2021, robeer-etal-2021-generating-realistic, wu-etal-2021-polyjuice, Gat2024}, and \textit{Matching methods},~\citep{Veitch2019Adapting, matching, Gat2024, Zhang2025MTCR}, and \textit{Concept Bottleneck models} \citep{Koh2020, Dalvi2022, Yu2024LACOAT}. Using LIBERTy, we evaluate representative methods from those approaches. Nevertheless, despite the advantages of concept-based explainability, particularly for end-users and decision-makers, it remains underexplored relative to token-level approaches \citep{CalderonReichart2024}. A possible reason for this gap is the current lack of benchmarks that enable rigorous evaluation and systematic comparison.

\paragraph{Explainability Benchmarks}~Benchmarking explanations is a highly challenging task, primarily because ground-truth explanations are rarely available in real-world datasets \citep{yang2019evaluating,hedstroem2023meta, XAI-Units, Seth2025bridging}. Most prior evaluation methods relied on indirect proxies, such as checking whether different methods agree with one another or whether their outputs align with simple heuristics \citep{hase2020evaluating,samek2021explaining}.
Furthermore, most explainability evaluations have focused on token-level explanations rather than reasoning over high-level semantic concepts \citep{thorne2019generating, wang2022fine, 2023rationalization}. As mentioned earlier, \textit{CEBaB} (Concept Effect Benchmark for NLP, \citet{cebab2022}) was the first dataset to evaluate explainability methods under controlled interventions. CEBaB revealed that many popular methods fail to estimate causal effects accurately and often perform no better than a naive concept-based matching baseline.

\begin{figure*}[!t]
\begingroup
\setlength{\jot}{2pt}
\setlength{\abovedisplayskip}{4pt}
\setlength{\belowdisplayskip}{6pt}
\setlength{\abovedisplayshortskip}{2pt}
\setlength{\belowdisplayshortskip}{2pt}
\begin{defprompt}[label={box:def_icace}]{CadetBlue}{Definitions: Causal Concept Effects and Estimators}
\begin{definition}[Causal Concept Effect (CaCE) and Individual CaCE (ICaCE)]
\[
\begin{aligned}
\mathrm{CaCE}_f(\stackrel{\rightarrow}{c}) 
  &= \E\left[f(X)\mid do(C=c')\right] 
    - \E\left[f(X)\mid do(C=c)\right] \\
\mathrm{ICaCE}_f(x_\varepsilon, \stackrel{\rightarrow}{c}) 
  &= \E\left[f(X)\mid do(C=c', \mathcal{E}=\varepsilon)\right] 
    - f\left(x_\varepsilon\right)
\end{aligned}
\]
\label{def:icace}
\end{definition}
\vspace{-1em}
\begin{definition}[Empirical CaCE and ICaCE]
\[
\begin{aligned}
\widehat{\mathrm{CaCE}}_f\left(\stackrel{\rightarrow}{c}\right)
  &= \frac{1}{|D|}\sum_{x_{\varepsilon^\ast}\in D}\Big[{f\left(\tilde{x}^{c^\ast\rightarrow c'}_{\varepsilon^\ast}\right)
          - f\left(\tilde{x}^{c^\ast\rightarrow c}_{\varepsilon^\ast}\right)}\Big] \\
\widehat{\mathrm{ICaCE}}_f\left(x_\varepsilon, \stackrel{\rightarrow}{c}\right)
  =& f\left(\tilde{x}^{\stackrel{\rightarrow}{c}}_\varepsilon\right) - f\left(x_\varepsilon\right)
\end{aligned}
\]
\label{def:estimators}
\vspace{-1em}
\end{definition}
\end{defprompt}
\endgroup
\vspace{-1.5em}
\end{figure*}

\citet{chaleshtori2024evaluating} recently noted an increasing need for richer benchmarks that capture the structural complexity of real-world data and enable the evaluation of both direct and indirect causal effects. Complementing this, \citet{Du2025IceCream} demonstrated that even state-of-the-art LLMs frequently fall prey to classical statistical fallacies, underscoring the limitations of existing evaluation methods in assessing true causal reasoning. We believe LIBERTy addresses these gaps by simulating realistic scenarios with diverse text types and rich causal graphs.


\section{Evaluation of Explanations}
\label{sec:evaluation}

In this section, we provide the relevant causal background and outline our causal approach to evaluating explanations of different scopes. \textit{Local explanations} capture how a concept influences a model's prediction for a specific instance, whereas \textit{global explanations} capture its influence across the entire data distribution. We evaluate explanations by comparing them with causal effects: local explanations against individual-level effects and global explanations against population-level effects.

\subsection{Causality Background}
\label{sec:causaility}

\paragraph{Structural Causal Models} 
We adopt the \textit{Structural Causal Model (SCM)} framework of \citet{pearl2009causality}. An SCM consists of exogenous and endogenous variables, together with structural equations.
Each endogenous variable is defined as a function of its parent endogenous variables and its associated exogenous noise variable. The induced causal graph is a directed acyclic graph encoding these dependencies. An example of a causal graph is given in Figure~\ref{fig:pipeline}. In this figure, the endogenous variables are the concepts ($A, B, C, Y$), the LLM-generated text $x_\varepsilon$, and the prediction of the explained model $f(x_\varepsilon)$. The exogenous variables include Gaussian noise terms ($\varepsilon_a, \varepsilon_b, \varepsilon_c, \varepsilon_y$), or randomly sampled auxiliary text provided to the LLM ($\varepsilon_{template}$ and $\varepsilon_{persona}$). Complementing the SCM with explicit distributions over the exogenous variables yields the \textit{data-generating process (DGP)}.

\paragraph{Counterfactuals} Within the SCM framework, a \textit{counterfactual} is the outcome of an intervention that assigns a different value to a concept while keeping all the exogenous variables fixed. In our setting, counterfactuals arise at two levels. First, a \textit{textual counterfactual} is generated by propagating the intervened concept assignment through the SCM. Second, a \textit{prediction counterfactual} is obtained by passing this counterfactual text to the explained model and observing its new prediction. Because the DGP is fully specified and LLM decoding is deterministic (with the temperature set to zero), the counterfactuals align with Pearl's definition of structural counterfactuals \citep{pearl2013structural}.

\paragraph{Causal Effect of Concepts} We consider two levels of causal effects: the \textit{Causal Concept Effect (CaCE)} \citep{goyal2019}, analogous to an Average Treatment Effect, and the \textit{individual CaCE (ICaCE)}, analogous to an Individual Treatment Effect, where the treatment is a concept, and the outcome is the model prediction. Ideally, a faithful explanation method would estimate the \textbf{CaCE as a global explanation} and the \textbf{ICaCE as a local explanation} \citep{Gat2024}. Formally, let $C$ denote the concept whose value changes from $c$ to $c'$ (written $\stackrel{\rightarrow}{c}$), and let $\mathcal{E}$ be exogenous variables with $\varepsilon$ values. Then, $x_\varepsilon$ is the resulting text , and the prediction of the explained model is $f(x_\varepsilon)$, which is a vector with softmax probabilities of each class of the concept $Y$ the model $f$ predicts. We denote expectations under the interventional distribution by the standard do-operator notation $\E\left[\cdot|do(C=c')\right]$ \citep{pearl2009causality}. The formal definitions are provided in Box~\ref{box:def_icace} Def~\ref{def:icace}. Both CaCE and ICaCE are vectors, capturing effects on all classes of $Y$.

\begin{figure*}[t]
\begingroup
\setlength{\jot}{2pt}
\setlength{\abovedisplayskip}{4pt}
\setlength{\belowdisplayskip}{5pt}
\setlength{\abovedisplayshortskip}{2pt}
\setlength{\belowdisplayshortskip}{2pt}
\begin{defprompt}[label={box:def_measures}]{CadetBlue}{Definitions: Evaluation Measures}
\begin{definition}[ICaCE Error Distance (ED)]
\vspace{-0.25em}
\[
\begin{aligned}
    \mathrm{ED}\left(f, M_{f}, x_\varepsilon, \stackrel{\rightarrow}{c}\right) &= \text{dist}\left( \widehat{\mathrm{ICaCE}}_f(x_\varepsilon, \stackrel{\rightarrow}{c})\textbf{; }  M_{f}(x_\varepsilon, \stackrel{\rightarrow}{c}) \right) \\
    \overline{\mathrm{ED}}\left(f, M_{f}\right) &= \frac{1}{|\mathcal{C}|}\sum_{\stackrel{\rightarrow}{c} \in \mathcal{C}}{\frac{1}{|D_{\stackrel{\rightarrow}{c}}|}\sum_{x_\varepsilon \in D_{\stackrel{\rightarrow}{c}}}{\mathrm{ED}\left(f, M_{f}, x_\varepsilon, \stackrel{\rightarrow}{c}\right)}}
\end{aligned}
\]
\label{def:ed}
\end{definition}
\vspace{-2em}
\begin{definition}[ICaCE Order-Faithfulness (OF)]
\[
\begin{aligned}
\mathrm{OF}\left(f, M_{f}, x_\varepsilon, \stackrel{\rightarrow}{c_1}, \stackrel{\rightarrow}{c_2}\right) 
    &= \text{sign}\Big(\widehat{\mathrm{ICaCE}}_f(x_\varepsilon, \stackrel{\rightarrow}{c_1}) - \widehat{\mathrm{ICaCE}}_f(x_\varepsilon, \stackrel{\rightarrow}{c_2})\textbf{; } M_{f}(x_\varepsilon, \stackrel{\rightarrow}{c_1}) - M_{f}(x_\varepsilon, \stackrel{\rightarrow}{c_2})\Big) \\
    \overline{\mathrm{OF}}\left(f, M_{f}\right) &= 
\frac{1}{|\mathcal{C}|(|\mathcal{C}|-1)} 
\sum_{\substack{\stackrel{\rightarrow}{c_1},\stackrel{\rightarrow}{c_2} \in \mathcal{C} \\ c_1 \neq c_2}}
\frac{1}{|D_{\stackrel{\rightarrow}{c_1}} \cap D_{\stackrel{\rightarrow}{c_2}}|}
\sum_{x_\varepsilon \in D_{\stackrel{\rightarrow}{c_1}} \cap D_{\stackrel{\rightarrow}{c_2}}}
\mathrm{OF}\left(f, M_{f}, x_\varepsilon, \stackrel{\rightarrow}{c_1}, \stackrel{\rightarrow}{c_2}\right)
\end{aligned}
\]
\label{def:of}
\vspace{-1.25em}
\end{definition}
Where ${dist}(\cdot ;\cdot)$ is a distance metric and ${sign}(\cdot;\cdot)$ is the proportion of vector entries that agree in sign. 
\end{defprompt}
\vspace{-1.5em}
\endgroup
\end{figure*}

\subsection{Evaluating Explanations}
\label{sub:evaluating}

\paragraph{Estimating Causal Effects} Both the CaCE and ICaCE are theoretical quantities, and in practice, we estimate them using counterfactuals. For $x_\varepsilon$, we denote its counterfactual by $\tilde{x}^{\stackrel{\rightarrow}{c}}_\varepsilon$. The formal definitions of the estimators are provided in Box~\ref{box:def_icace} Def~\ref{def:estimators}. In our setting, $\widehat{\mathrm{ICaCE}}_f$ is exact because, with fixed $\mathcal{E}$ and deterministic decoding, 
$\mathbb{E}\left[f(X)\,\middle|\, \mathrm{do}(C{=}c',\, \mathcal{E}{=}\varepsilon)\right]
= f\left(\tilde{x}^{\stackrel{\rightarrow}{c}}_\varepsilon\right)$.
If decoding is stochastic (e.g., temperature $>0$), additional noise is introduced through token sampling (see the discussion in Appendix~\ref{sec:discussion}).

\paragraph{Evaluation Pipeline} 
The explained model $f$ is trained on DGP-sampled data $D_f$.\footnote{No training is required if $f$ is a zero/few-shot LLM.}
The explanation method $M$ is trained on pairs ${(x,f(x)) : x\in D_M}$, with optional access to gold concept values or other auxiliary information, depending on the evaluator choice. 
For evaluating $M_f$, we use the interventional test set $D_\mathcal{C}$, where $\mathcal{C}$ denotes the set of concept changes. For each change, $D_{\stackrel{\rightarrow}{c}}$ consists of pairs of textual examples and their counterfactuals, $(x_{\varepsilon}, \tilde{x}^{\stackrel{\rightarrow}{c}}_\varepsilon)$. From these, we compute $\widehat{\mathrm{CaCE}}_f(\stackrel{\rightarrow}{c})$ and $\left\{\widehat{\mathrm{ICaCE}}_f(x, \stackrel{\rightarrow}{c})\right\}_{x}$, as well as the corresponding explanation scores: $M_{f}(\stackrel{\rightarrow}{c})$ for global methods and $\left\{M_{f}(x, \stackrel{\rightarrow}{c})\right\}_{x}$ for local ones. We next describe the evaluation measures for local explanations, which can be extended to global explanations with minor modifications.

\paragraph{Evaluation Measures} CEBaB reports the average ICaCE Error Distance over all the concept changes, defined as the distance between the reference effects and the explanation (formal definition in Box~\ref{box:def_measures} Def~\ref{def:ed}). Following \citet{cebab2022}, we consider three distance metrics: cosine distance, L2 distance, and norm difference. We use their mean as the final reported error distance ($\overline{\mathrm{ED}}$).

In addition, we propose a new measure, which we call \textit{Order-Faithfulness}. 
This measure builds on the necessary condition for faithful explanations introduced by \citet{Gat2024}, 
which states that an explanation must rank one concept as more important than another if and only if its true causal effect is larger. While ED measures estimation accuracy, Order-Faithfulness assesses whether explanations preserve the relative ordering of concept importance, a property that is often more robust, interpretable, and directly relevant to how explanations are used in practice. To formalize this idea, consider two concept changes $\stackrel{\rightarrow}{c_1}$ and $\stackrel{\rightarrow}{c_2}$. We first compute the difference between their reference effect vectors, 
and then the difference between their explanation vectors. 
We compare the signs of each entry in the difference vector with the corresponding entry in the explanation difference vector. Agreement of signs indicates that the explanation preserves the correct ordering of the two concept changes, and is therefore \textit{order-faithful}.
The formal definition is provided in Box~\ref{box:def_measures} Def~\ref{def:of}. To summarize, we report the average error distance $\overline{\mathrm{ED}}$ (lower is better) and the average order-faithfulness $\overline{\mathrm{OF}}$ (higher is better) to compare explanation methods. 
\section{Interventional Data Generation}
\label{sec:Causal_graph}

We next describe the process for generating an interventional benchmark using LIBERTy (\textbf{L}LM-based \textbf{I}nterventional \textbf{B}enchmark for \textbf{E}xplainability with \textbf{R}eference \textbf{T}argets). The framework relies on explicitly defined DGPs comprising three components: SCMs over concepts, exogenous grounding texts, and an LLM (see Figure~\ref{fig:pipeline}). These DGPs allow us to generate silver counterfactuals.

\begin{figure*}[t]
    \centering
    \includegraphics[width=0.98\textwidth]{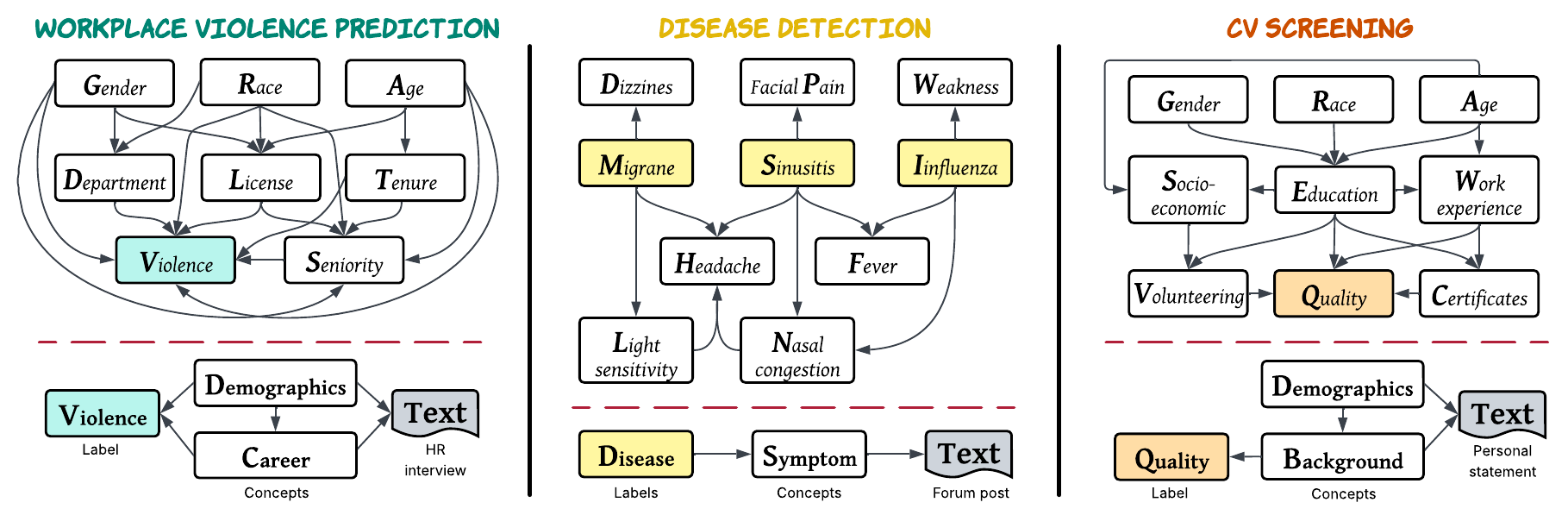}
    \caption{\textbf{LIBERTy Causal Graphs:} We show only the concepts (endogenous variables) and the relationships between them. Colored concepts indicate the variables that the explained model is trained to predict (the $Y$). At the bottom, a simplified version is provided. The graphs are grounded in prior literature and studies.}
    \label{fig:causal_graphs}
\vspace{-0.8em}
\end{figure*}

\paragraph{LIBERTy SCMs}~
For each dataset, we first define the causal graph that specifies the concepts and their directional relationships (which concept influences which). 
Based on this graph, we specify the structural equations: each concept is linked to a function that determines its value based on its parent concepts and an exogenous noise term. The noise is drawn from a Gaussian distribution, with a concept-specific mean and variance. 
The structural equation generating the text takes all concepts as inputs and uses two exogenous grounding texts, a persona and a template, instead of Gaussian noise.

In Figure~\ref{fig:causal_graphs}, we illustrate the three causal graphs of the three LIBERTy datasets. 
While their SCMs are not intended to mirror the true causal structure of the world (see the discussion in Appendix~\ref{sub:real_world}), they are grounded in plausible assumptions: one causal graph (workplace violence prediction) is adapted from prior literature \citep{violence_nurse_study}, and the other two (disease detection and CV screening) are informed by statistical patterns in real-world data \citep{monto2000clinical, cady2002sinus, dastin2018amazon}. 
Finally, we note that our three causal graphs are much more complex and richer than the (four-concepts) causal graph of CEBaB. Each graph includes at least eight concepts, exhibits confounding and mediation structures (allowing estimation of direct and indirect effects), contains long paths (up to four edges between a concept and the text), and supports both anticausal ($Y\!\rightarrow\!Text$) and confounded ($Y\!\leftarrow\!C\!\rightarrow\!Text$) learning problems.

\paragraph{Exogenous Grounding Texts}~
To ensure the validity of our structural counterfactuals, deterministic decoding is required. With stochastic decoding, generation noise cannot be tracked or held fixed across factual and counterfactual texts, causing them to differ in unobserved exogenous factors rather than only in the intervened concepts, and thus violating the definition of a structural counterfactual (see Appendix~\ref{sub:deterministic}). However, this requirement introduces its own limitations. First, for a given combination of concept values, deterministic decoding produces a single, fixed text. Second, this decoding yields highly generic, templated, and repetitive texts, regardless of concept values (always the same narrative, albeit with minor variations). Third, the generated examples do not seem like authentic human-written text. To address these limitations, we propose a simple yet elegant solution in the spirit of the SCM framework: we introduce two additional exogenous variables, an author persona and a template, both of which serve as a grounding context for the LLM. 

The \textit{Persona variable} $\varepsilon_{persona}$ represents a set of contextual attributes, including profession, hobbies, and personal motivations. In contrast, the \textit{Template variable} $\varepsilon_{template}$ captures a particular discourse structure, derived from real-world corpora (e.g., personal statements, Reddit posts). Templates and personas support three key goals: (1) making the generated texts resemble authentic text; (2) promoting diversity: for each set of concept values, there are $|\mathcal{E}_{persona}|\times|\mathcal{E}_{template}|$ possible instantiations; and (3) ensuring the original example and its counterfactual derive from the same narrative. 

\paragraph{Text Generation}~
We sample concept values in topological order from the SCM, using the equations and Gaussian noise. 
We then sample a persona and a template and record all variable values for later counterfactual generation.
Textual realizations are generated via deterministic decoding (zero temperature) by conditioning GPT-4o on the full set of concept values, along with the persona and template. We use a dedicated prompt for each dataset. Notably, GPT-4o receives only the concept values and does not observe the causal graph itself.

\paragraph{Counterfactual Generation}~
We follow Pearl’s three-step counterfactual procedure \citep{pearl2009causality}.
(1) Abduction: fix the exogenous variables used for the original example,
(2) Action: intervene on a target concept, and
(3) Prediction: propagate the intervention through the SCM and compute updated concept values. 
We then regenerate the text using the same persona, template, and deterministic decoding. The red arrows in Figure~\ref{fig:pipeline} illustrate this. For each test example, we randomly select three concept changes and generate counterfactuals.
\begin{table}[t]
\centering
\normalsize
\begin{adjustbox}{width=0.45\textwidth}
\begin{tabular}{lcccc}
\toprule
\textbf{Dataset} & \textbf{$D_{\stackrel{\rightarrow}{c}}$} & \textbf{Pairs} & \textbf{Words} \\
\midrule
Workplace Violence     &  1756 & 1317 & 350.9 \\
Disease Detection  & 1243 & 932  & 310.8 \\
CV Screening       &  1332 & 998  & 313.0 \\
\bottomrule
\end{tabular}
\end{adjustbox}
\caption{\textbf{Data Statistics:} For all datasets, $|D_f|$=1.5K and $|D_M|$=0.5K. \textbf{Pairs} is the number of $(x_{\varepsilon}, \tilde{x}^{\stackrel{\rightarrow}{c}}_\varepsilon)$ pairs in $D_{\stackrel{\rightarrow}{c}}$. \textbf{Words} reports the average number per example.}
\label{tab:stats}
\vspace{-0.8em}
\end{table}

\section{Datasets}
\label{sec:datasets}

LIBERTy comprises three datasets, each modeling a high-stakes, socially impactful NLP task where explainability is critical. Each dataset is divided into four subsets: two for training and testing the explained model, one for training the explanation method, and one test set containing pairs of texts and their counterfactuals. The first three subsets exclude counterfactuals, which are unavailable in real-world settings.
The number of examples in each dataset is provided in Table~\ref{tab:stats}.
The LLM integrated within the SCMs (for generating texts) is GPT-4o, while Gemini-1.5-Pro is used to create templates and personas. Below, we briefly describe each dataset. Due to space limitations, the SCMs, prompts, representative examples, and additional technical details are provided in Appendix~\ref{sec:dataset_details}.

\subsection{Workplace Violence Prediction}
This dataset simulates HR–nurse interviews, in which the (explained) model predicts the likelihood that a nurse will experience workplace violence. The causal graph is adapted from the Minnesota Nurses' Study \citep{violence_nurse_study}, which documented the prevalence of verbal and physical violence among clinical staff and analyzed risk factors by demographic and professional background. The template follows a structured HR interview format. To ensure both realism and sufficient diversity, we generate interview templates as follows: for each concept, a bank of 10 questions is created using Gemini, each designed to elicit the concept's value from different linguistic perspectives. Additionally, 10 opening and 10 closing sentence variants are defined to maintain a coherent interview flow. Each template is generated by sampling one question per concept, along with an opening and closing sentence. The question order is randomized, yielding a large pool of interview templates. The persona contains three informal ``fun facts'' about the nurse, each centered on a concept (without specifying its value). Using Gemini, we generated 500 personas. Additional details are in Appendix~\ref{sub:workplace_violence}.

\subsection{Disease detection}
This dataset simulates clinical self-reports, where the (explained) model predicts a disease from symptoms described in a medical forum post. Unlike the other two datasets, the learning problem is anti-causal: the disease label serves as the root cause in the SCM and determines the values of symptom concepts, based on known symptom–disease relations \citep{monto2000clinical, cady2002sinus}. The template is a narrative structure abstracted from 1,310 posts on Reddit's DiagnoseMe forum,\footnote{\url{https://www.reddit.com/r/DiagnoseMe/}} using Gemini to preserve the clinical tone and flow. The persona (a total of 1200) consists of three informal facts about occupation, hobbies, and family or friends. To generate personas, we first sample an occupation and a hobby from predefined lists, then use Gemini to generate the corresponding facts. Each dataset example is created by prompting GPT-4o to follow the template and integrate information from the persona and the symptom values. 
Additional details are in Appendix~\ref{sub:disease_detection}.

\subsection{CV Screening}

This dataset simulates automated resume assessment, where the model is tasked with predicting an applicant's quality from a CV-style personal statement, with labels such as weak, qualified, and outstanding. Motivated by critiques of real-world screening systems \citep{dastin2018amazon, raghavan2020mitigating, cowgill2021biased}, the causal graph encodes hypothesized dependencies between demographic and professional attributes, inspired by statistical patterns reported by the U.S. Bureau of Labor Statistics.\footnote{\url{https://www.bls.gov/cps/demographics.htm}} For example, gender influences the hiring label only indirectly through mediators such as education and Work Experience.
1,235 templates were generated from 342 scraped personal statement examples,\footnote{\url{https://universitycompare.com}}
where each source text was abstracted with Gemini using a 2-shot prompt to produce several occupation-agnostic variants that preserve the narrative structure while removing concept- and role-specific details.
To generate a persona (a total of 990), we sample a role from a predefined list and use Gemini with a 2-shot prompt to produce both personal and professional context, including motivations and skills relevant to that role. Each dataset example is then created by prompting GPT-4o to follow the template and integrate information from the application role, the persona, and the sampled concept values. 
Additional details are in Appendix~\ref{sub:cv_screening}.
\begin{table*}[!t]
\centering
\footnotesize
\begin{adjustbox}{width=0.95\textwidth}
\begin{tabular}{l|cc|cccccc|cccccc}
\toprule
& \cellcolor{Gray!40} & \cellcolor{Gray!40} & \multicolumn{6}{c|}{\cellcolor{Gray!40}\textbf{Dataset}} & \multicolumn{6}{c}{\cellcolor{Gray!40}\textbf{Explained Model}} \\
 & \multicolumn{2}{c|}{\cellcolor{Gray!40}\multirow{-2}{*}{{\textbf{Average}}}} &
\multicolumn{2}{c}{\cellcolor{Gray!15}\textbf{Violence}} &
\multicolumn{2}{c}{\cellcolor{Gray!15}\textbf{Disease}} &
\multicolumn{2}{c|}{\cellcolor{Gray!15}\textbf{CV}} &
\multicolumn{2}{c}{\cellcolor{Gray!15}\textbf{DeBERTa-v3}} &
\multicolumn{2}{c}{\cellcolor{Gray!15}\textbf{Qwen-2.5}} &
\multicolumn{2}{c}{\cellcolor{Gray!15}\textbf{GPT-4o}} 
\\
$\downarrow$ \textbf{Method} & \cellcolor{Gray!15}ED & \cellcolor{Gray!15}OF & \cellcolor{Gray!15}ED & \cellcolor{Gray!15}OF & \cellcolor{Gray!15}ED & \cellcolor{Gray!15}OF & \cellcolor{Gray!15}ED & \cellcolor{Gray!15}OF & \cellcolor{Gray!15}ED & \cellcolor{Gray!15}OF & \cellcolor{Gray!15}ED & \cellcolor{Gray!15}OF & \cellcolor{Gray!15}ED & \cellcolor{Gray!15}OF \\
\midrule
\textit{CF Gen}   & 0.55 & 0.49 & 0.47 & 0.58 & 0.67 & 0.36 & 0.52 & 0.52 & 0.50 & 0.59 & 0.62 & 0.53 & 0.58 & 0.49 \\
\midrule
\textit{Approx}   & 0.45 & 0.69 & 0.41 & 0.71 & 0.48 & 0.69 & 0.46 & 0.66 & 0.38 & 0.76 & 0.50 & 0.70 & 0.53 & 0.67 \\
\textit{ConVecs}  & 0.44 & 0.69 & 0.40 & 0.73 & 0.44 & 0.70 & 0.47 & 0.66 & 0.34 & 0.78 & 0.47 & 0.71 & 0.52 & 0.68 \\
\textit{ST Match} & 0.49 & 0.65 & 0.51 & 0.63 & 0.46 & 0.69 & 0.50 & 0.62 & 0.49 & 0.69 & 0.55 & 0.66 & 0.53 & 0.67 \\
\textit{PT Match} & 0.51 & 0.64 & 0.51 & 0.64 & 0.52 & 0.65 & 0.50 & 0.63 & 0.52 & 0.68 & 0.56 & 0.65 & 0.59 & 0.64 \\
\textit{FT Match} & \textbf{0.34} & \textbf{0.74} & \textbf{0.32} & \textbf{0.76} & \textbf{0.36} & \textbf{0.75} & \textbf{0.35} & \textbf{0.72} & \textbf{0.16} & \textbf{0.88} & \textbf{0.39} & \textbf{0.75} & \textbf{0.48} & \textbf{0.70} \\
\midrule
\textit{LEACE}    & 0.65 & 0.46 & \textemdash & \textemdash & 0.65 & 0.46 & \textemdash & \textemdash & 0.62 & 0.42 & 0.87 & 0.41 & \textemdash & \textemdash \\
\bottomrule
\end{tabular}
\end{adjustbox}
\caption{\textbf{Local Explainability Results:} We report the Average ICaCE Error-Distance ($\overline{\mathrm{ED}}\ge0$, $\downarrow$ is better) and Average ICaCE Order-Faithfulness ($\overline{\mathrm{OF}}\le1$, $\uparrow$ is better). The \textbf{Average} column reports the mean across five explained models and three datasets. 
The detailed results appear in Appendix Table~\ref{tab:complete_local} and exhibit a similar pattern, with fine-tuned matching outperforming other approaches.
Horizontal lines separate method families.}
\label{tab:local_results}
\vspace{-0.8em}
\end{table*}

\section{Experimental Setup}

Using LIBERTy, we conduct experiments on five explained models and benchmark eight explanation methods from four families of approaches. The goals of our experiments are:
(1) Benchmarking local and global explanation methods;
(2) Analyzing the sensitivity of models to concept changes and evaluating which model captures better the causal structure of the data. The evaluation pipeline is described in Section~\ref{sub:evaluating}. When reporting scores, we typically average them over all concept changes.

\paragraph{Explained Models}~We evaluate five models. Three are fine-tuned to predict $Y$ from text: (1) DeBERTa-v3 (base, \citet{deberta}), an encoder-only model; (2) T5 (base, \citet{t5}), an encoder–decoder model;
 and (3) Qwen-2.5 (1.5B-instruct, \citet{qwen}), a decoder-only LLM. The other two are zero-shot LLMs: (4) Llama-3.1 (8B-instruct, \citet{llama3}) and (5) GPT-4o \citep{gpt4o}. See Appendix~\ref{sub:explained} for more details, hyperparameters, performance, and prompts.

\paragraph{Explainability Methods}~We briefly mention the explainability methods we benchmark, but Appendix~\ref{sec:explainability} thoroughly describes and discusses them. The rationale for selecting methods was to focus on top-performing approaches previously applied to CEBaB with user-friendly code. We examine eight methods covering four families: 

(1) \textit{Counterfactual Generation}: LLMs generate counterfactuals by editing texts to reflect a target concept change \citep{Gat2024}. We examine in Appendix~\ref{sub:counterfactuals} four prompting techniques, each injecting different causal assumptions. We mainly focus on the \textit{Mediators and Confounders} technique, which fixes confounders while allowing mediators to vary, and achieves the best performance.

(2) \textit{Matching} (see Appendix~\ref{sub:matching}): matching methods search for the most similar candidate from a predefined set of examples with the target concept change. The difference between the methods lies in how similarity is defined, and we examine five methods: (2a) \textit{ST Match}: cosine similarity over SentenceTransformer embeddings \citep{ReimersG19}; (2b) \textit{PT Match}: cosine similarity over a pre-trained encoder-only model (DeBERTa); (2c) \textit{FT Match}: cosine similarity over an encoder fine-tuned to predict $Y$; (2d) \textit{Approx}: first predicts concept values using fine-tuned models and then search for exact concept-based match; and (2e) \textit{ConVecs}: cosine similarity over concatenated softmax prediction vectors of all concepts. Notably, the first three are semantic-based methods, while the latter two are concept-based ones.

(3) \textit{Concept Erasure} (see Appendix~\ref{sub:concept_erasure}): removes linearly encoded information about a target concept from hidden representations using \textit{LEACE} \citep{belrose2023leace}.\footnote{We employ \textit{LEACE} only for open-source models and only on the Disease Detection dataset, where erasing a concept is well defined as its absence (e.g., symptom not present).}
(4) \textit{Concept Attributions} (see Appendix~\ref{sub:attributions}): estimates concept importance via \textit{ConceptShap} \citep{Human_KeyConcepts} combined with \textit{TCAV} \citep{TCAV_Kim2018}, which construct concept vectors and assign Shapley-based scores.\footnote{We benchmark \textit{ConceptShap} only as a global explanation for open-source models.}
\section{Results}

\subsection{Local Explanations}

We begin by comparing the local explainability methods using LIBERTy, reporting ICaCE $\overline{\text{ED}}$ and $\overline{\text{OF}}$. Table~\ref{tab:local_results} presents these measures at the dataset level (averaged across all five models) and at the model level (averaged across all three datasets). Complete results are provided in Table~\ref{tab:complete_local} (Appendix~\ref{sec:additional}). Overall, the matching approach performs best. Within this category, \textit{FT Match}, which fine-tunes an encoder-only model to predict the label $Y$ and then uses its embeddings for similarity, achieves the lowest estimation error and emerges as the most faithful method.
Its advantage likely stems from the model learning task-specific representations that produce more meaningful neighborhoods for matching. 
Other strong performers are the concept-based matching methods, \textit{ConVecs} (proposed in this work) and \textit{Approx}. These findings align with those of \citet{Gat2024}, who compared different matching methods on CEBaB and reported similar trends.

\begin{figure}[t]
    \centering
    \includegraphics[width=0.425\textwidth]{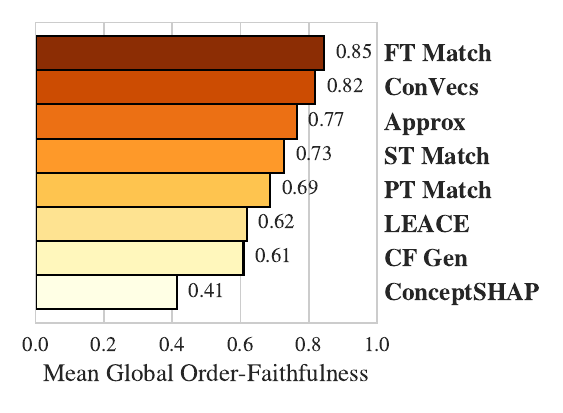}
    \vspace{-1em}
    \caption{\textbf{Global Explainability Results:} We report the mean Order-Faithfulness score for global explanations. See Table~\ref{tab:complete_global} in the Appendix for full results.}
    \label{fig:global_exp}
\vspace{-0.8em}
\end{figure}
\begin{table}[t]
\centering
\footnotesize
\begin{adjustbox}{width=0.48\textwidth}
\begin{tabular}{l|c|c|c}
\toprule

\cellcolor{Gray!40} \textbf{Dataset} & \cellcolor{Gray!40}
\textbf{Violence} & \cellcolor{Gray!40} \textbf{Disease} & \cellcolor{Gray!40} \textbf{CV} \\ 
\cellcolor{Gray!15} Model & \cellcolor{Gray!15} Qwen-2.5 & \cellcolor{Gray!15} DeBERTa-v3 
 & \cellcolor{Gray!15} GPT-4o \\
\midrule
\multirow{3}{*}{Gold} & \cellcolor{RoyalBlue!50} Gender & \cellcolor{Green!50} Light Sens & \cellcolor{BrickRed!50} Work Exp \\
 & \cellcolor{TealBlue!50} Department & \cellcolor{SpringGreen!50} Facial Pain & \cellcolor{Orange!50} Education \\
 & \cellcolor{Turquoise!15} Age & \cellcolor{Goldenrod!50} Dizziness & \cellcolor{Apricot!50} Race \\
\midrule
\multirow{3}{*}{FT Match} & \cellcolor{RoyalBlue!50} Gender & \cellcolor{Green!50} Light Sens & \cellcolor{Orange!50} Education \\
 & Seniority & \cellcolor{Goldenrod!50} Dizziness & \cellcolor{BrickRed!50} Work Exp \\
 & \cellcolor{Turquoise!15} Age & \cellcolor{SpringGreen!50} Facial Pain & Age \\
\midrule
\multirow{3}{*}{CF Gen} & \cellcolor{RoyalBlue!50} Gender & Weakness & \cellcolor{Orange!50} Education \\
 & \cellcolor{Turquoise!15} Age & \cellcolor{Goldenrod!50} Dizziness & \cellcolor{BrickRed!50} Work Exp \\
 & Race & \cellcolor{Green!50} Light Sens & Socioeco \\
\midrule
\multirow{3}{*}{LEACE} &  & \cellcolor{Goldenrod!50} Dizziness &  \\
 &  & \cellcolor{Green!50} Light Sens &  \\
 &  & Headache &  \\
\midrule
\multirow{3}{*}{ConceptShap} & \cellcolor{RoyalBlue!50} Gender & \cellcolor{Goldenrod!50} Dizziness &  \\
 & Race & Nasal Cong &  \\
 & Seniority & Weakness &  \\
\bottomrule
\end{tabular}
\end{adjustbox}
\caption{\textbf{Global Explanations Analysis:} We present the top-3 most important concepts of explanations for selected datasets, models, and methods. A colored concept indicates it is among the top three gold concepts.}
\label{tab:global_analysis}
\vspace{-0.8em}
\end{table}

An interesting difference between our findings and those of \citet{Gat2024} is that, while LLM-generated counterfactuals outperform matching-based methods on CEBaB, the opposite holds on LIBERTy.
A potential explanation is that humans write CEBaB's counterfactuals: annotators edit an existing text to reflect a change in concept. LLMs can closely mimic this editing process, especially for short, simple texts, which makes their generated counterfactuals appear effective. In LIBERTy, producing an explanation that resembles a human edit does not guarantee faithfulness; instead, the explanation should reflect the actual DGP. This also explains why matching methods perform more consistently: their retrieved candidates are sampled from distributions aligned with the underlying DGP rather than produced through human-aligned textual edits. We refer the reader to an extended discussion of these aspects in Appendix~\ref{sub:counterfactual_fails}.

Finally, the $\overline{\text{ED}}$ and $\overline{\text{OF}}$ scores reveal substantial room for improvement. In LIBERTy, even the best methods achieve only around 0.3 on ED (where 0 is perfect) and 0.7 on OF (where 1 is perfect). We hope that LIBERTy will encourage further progress on developing more faithful explanation methods.

\begin{table*}[t]
\centering
\footnotesize
\begin{adjustbox}{width=0.98\textwidth}
\begin{tabular}{l|ccc|cc|ccc}
\toprule
 \cellcolor{Gray!15} \textbf{Examined} & \multicolumn{3}{c}{\cellcolor{Gray!40}\textbf{Workplace Violence}} & \multicolumn{2}{c}{\cellcolor{Gray!40}\textbf{Disease Detection}} & \multicolumn{3}{c}{\cellcolor{Gray!40}\textbf{CV Screening}} \\
\cellcolor{Gray!15} 
\textbf{Model} & \cellcolor{Gray!15} Race & \cellcolor{Gray!15} Gender & \cellcolor{Gray!15} Age & \cellcolor{Gray!15} Headache & \cellcolor{Gray!15} General Weakness & \cellcolor{Gray!15} Race & \cellcolor{Gray!15} Gender & \cellcolor{Gray!15} Age  \\
\midrule
DeBERTa-v3 & 0.350 & 1.192 & 0.758 & 0.398 & 0.415 & \textbf{0.715} & 0.432 & \textbf{0.613} \\
T5 & \textbf{0.421} & 0.743 & 0.512 & 0.530 & 0.376 & 0.742 & 0.398 & 0.513 \\
Qwen-2.5 & 0.691 & \textbf{1.314} & \textbf{1.045} & 0.426 & 0.512 & 0.522 & \textbf{0.361} & 0.503 \\
Llama-3.1 & 0.224 & 0.227 & 0.226 & 0.364 & 0.332 & 0.374 & 0.283 & 0.397 \\
GPT-4o & 0.724 & 0.594 & 0.300 & 0.369 & 0.215 & 0.417 & 0.208 & 0.355 \\
\cellcolor{Goldenrod!80} True Effect & \cellcolor{Goldenrod!80} 0.484 & \cellcolor{Goldenrod!80} 1.271 & \cellcolor{Goldenrod!80} 1.154 & \cellcolor{Goldenrod!80} -- & \cellcolor{Goldenrod!80} -- & \cellcolor{Goldenrod!80} 0.636 & \cellcolor{Goldenrod!80} 0.369 & \cellcolor{Goldenrod!80} 0.913 \\
\bottomrule
\end{tabular}
\end{adjustbox}
\caption{\textbf{Concept Sensitivity Analysis:} 
In the Disease Detection dataset, $Y$ is the parent of the concepts, so interventions do not affect $Y$, and its ground-truth sensitivity cannot be computed.
 See Table~\ref{tab:complete_sensitivity} for full results.} 
\label{tab:sensitivity}
\vspace{-0.8em}
\end{table*}

\subsection{Global Explanations}

Many global explanations produce a ranked list of concepts by their overall importance (not specific to a single example), reflecting their influence on the model's predictions (a.k.a. feature importance). We therefore evaluate their global order-faithfulness: are the concepts ranked in the same order as their causal effects? To obtain the ground-truth ranking, we compute a single gold importance score for each concept using CaCE. Note that for each concept change, CaCE yields a vector of size $|Y|$, capturing the causal effect of that change on each output class. To obtain a single gold importance score for each concept, we first sum the absolute CaCE values across all output classes, reflecting the total magnitude of the effect for that change. We then average this quantity over all changes. This produces a single gold importance score per concept. Global $\overline{\text{OF}}$ is then computed based on these scores: it quantifies how faithfully each explanation method's ranking of concept importance matches the gold ranking.\footnote{Global $\overline{\text{OF}}$ and ICaCE $\overline{\text{OF}}$ differ both in the order of computation and in what is being ranked. ICaCE $\overline{\text{OF}}$ evaluates order-faithfulness over individual concept changes on a per-example basis before averaging, whereas Global $\overline{\text{OF}}$ evaluates order-faithfulness over concepts, using global importance scores derived from CaCE.}

Figure~\ref{fig:global_exp} compares the methods using the average global $\overline{\text{OF}}$ across the three datasets and five models. 
The complete (non-averaged) results are reported in Table~\ref{tab:complete_global} in the Appendix.
As shown, global trends mirror the local ones, with the matching approach outperforming the others. Table~\ref{tab:global_analysis} further reports the top-3 most important concepts identified by each method and compares them with the top-3 gold concepts (according to their gold importance score). Every method misses at least one gold concept, highlighting the need for further research on global explainability.

\subsection{Sensitivity Analysis}

Up to this point, we have used LIBERTy to evaluate explanation methods. More broadly, the framework supports two complementary analyses. First, it can be used to analyze a model's sensitivity to concept changes by measuring the magnitude of prediction changes induced by structural counterfactuals. Second, LIBERTy can be used to assess how well different learning methods, such as CE fine-tuning, align model behavior with the causal structure encoded in the DGP. This second analysis necessarily focuses on models trained on the generated data, since only then can their behavior be expected to reflect the underlying causal relationships. Under successful learning, a model's sensitivity to concept changes should closely match the true causal effects on the outcome variable $Y$, which we estimate via Monte Carlo simulation from the SCM.

For a given example and concept change, we compute a sensitivity score that quantifies the extent to which the model's prediction is affected. This score is obtained by summing the absolute ICaCE values, which quantify the magnitude of the change. Larger values indicate stronger shifts in the prediction (i.e., more sensitive). Table~\ref{tab:sensitivity} reports sensitivity scores for the five evaluated models on selected concepts (an average over all their changes), alongside the gold sensitivity effect.
Complete results are provided in Table~\ref{tab:complete_sensitivity} in the Appendix.

When examining sensitivity scores (without comparing them to the gold effects) we observe that zero-shot LLMs (Llama-3.1-8B and GPT-4o) exhibit lower sensitivity to concept changes, particularly for demographic concepts such as Race, Gender, and Age (Table~\ref{tab:sensitivity}). We believe the reduced sensitivity reflects intentional design choices made during post-training alignment. In addition, among the fine-tuned models, we find that Qwen2.5-1.5B most accurately reflects the causal structure of the data. Still, the gap with the gold effects highlights that fine-tuning is insufficient and that there remains a need for causal learning techniques.
\section{Conclusions}

A central challenge in explainability is the lack of reliable evaluation protocols, particularly given the absence of ``gold explanations''. Our work takes a significant step toward closing this gap. We introduced LIBERTy, a framework for generating interventional datasets to benchmark concept-based explanations against ``silver'' references: causal effects estimated using structural counterfactuals. Using LIBERTy, we evaluated local and global explainability methods, the sensitivity of LLMs to concept interventions, and the causal learning capabilities of fine-tuned models.

In Section~\ref{sub:opportunities} in the Appendix, we outline future research opportunities motivated by our four key findings. First, we found that LLM-generated counterfactuals, which were previously reported as state-of-the-art explanations \citep{Gat2024}, do not retain this status when evaluated against structural counterfactuals (as in LIBERTy) rather than human-written ones (as in CEBaB). This highlights the need for a broader evaluation of explanations. Second, we observed a large room for improvement in both local and global explanations, offering clear targets for future work. 

Third, our concept-sensitivity analysis showed that some LLMs are largely insensitive to demographic interventions, likely due to post-alignment mitigation effects. Finally, our analysis revealed that vanilla fine-tuning may fail to capture the causal structure of the data, suggesting the need for unique learning methods. To summarize, there is great promise in developing smaller, theory-grounded, causal-inspired explainability and learning approaches. We hope our work will serve as a foundation for such future research.
\section{Limitations}

\paragraph{Synthetic Text Generation}
LIBERTy relies on LLMs to instantiate structural counterfactuals. However, it also means that the texts are synthetic rather than human-written. This may introduce mismatches between how the LLM instantiates concepts and how humans would naturally express them.
To assess data quality, we conducted a human evaluation (Appendix~\ref{sec:human_val}). Annotators confirmed that the generated texts are coherent, relevant, and fluent; that the LLM correctly incorporates concept values; and that counterfactuals are perceived as realistic variants differing in only one concept. Finally, although LIBERTy uses synthetic text, this limitation is increasingly less restrictive: a growing share of real-world data is generated by LLMs, making synthetic settings both common and practically meaningful. It is therefore reasonable to assume that model inputs in many future applications will themselves be LLM-generated.

\paragraph{Focusing on Concept-based Explanations}
Our work focuses exclusively on concept-based explanations and their causal evaluation. This scope covers only a subset of existing explainability methods, and most prior work centers on token-level or free-text explanations (see the analysis of \citet{CalderonReichart2024}). Nevertheless, there are strong reasons to focus on concept-based methods. These methods quantify how high-level, human-interpretable concepts (a.k.a. attributes, features, variables, or rubrics) that are implicitly or explicitly expressed in text influence the model. Because such high-level concepts align with human cognitive processes \citep{Alqaraawi2020, Kim2023, Poeta2023}, reduce the complexity of long inputs, and communicate model behavior in intuitive terms \citep{CalderonReichart2024}, concept-based explanations are particularly suitable for high-stakes settings where end users and decision makers must understand and trust model reasoning. We believe that the relatively limited attention to concept-based explainability stems partly from the lack of appropriate benchmarks for developing and evaluating such methods. By providing an interventional benchmark with structural causal effects, LIBERTy aims to address this gap and facilitate broader research and adoption of concept-based explanations.

\paragraph{DGPs as Approximations of Reality}
LIBERTy provides structural counterfactuals in the strict sense, as they are generated from a fully specified data-generating process (DGP). While the DGP and causal graphs only simplify real-world mechanisms, they are not arbitrary and are grounded in domain knowledge and the literature. Still, we acknowledge that they do not perfectly mirror real-world causal structures. Crucially, this limitation does not compromise the reliability of our evaluation protocol, because our goal is not to recover real-world mechanisms or estimate real-world causal effects. Instead, our objective is to measure the causal effects \textit{within the explained model} and benchmark explanation methods against those effects. For this purpose, what matters is that the DGP supports precise interventions and produces structural counterfactuals that faithfully reflect them. Explanation faithfulness is always defined relative to the explained model, whether its behavior arises from true causal relationships, simplified abstractions, or even spurious correlations. Thus, a synthetic DGP is sufficient and, in practice, often required for the controlled and rigorous evaluation of explanation methods. Such methods can be trained on or applied to data generated by the DGP and evaluated against the explained model's predictions, whether or not the model itself was trained on that data. We do not claim that our benchmark explains real-world phenomena or reveals how LLMs internally represent them. Rather, our goal is to provide a principled benchmark for comparing explanation methods, analyzing their limitations, and identifying those that most faithfully capture model behavior, thereby enabling their application in real-world settings. Please also see our discussion in Appendix~\ref{sub:real_world}.

\section*{Acknowledgments}

\bibliography{custom}


\appendix

\renewcommand \thepart{}
\renewcommand \partname{}
\mtcsettitle{parttoc}{}
\addcontentsline{toc}{section}{Appendix} 
\part{Appendix} 
\parttoc 

\section{Discussion}
\label{sec:discussion}

\subsection{Real-World Data}
\label{sub:real_world}

\textbf{Why is it acceptable that the LIBERTy SCM does not perfectly reflect the real world?}~We do not claim that our benchmark explains real-world phenomena or reveals how LLMs internally represent them. Rather, our goal is to provide a principled benchmark for comparing explanation methods, analyzing their limitations, and identifying those that most faithfully capture model behavior, thereby enabling their application in real-world settings. Therefore, it does not matter whether LIBERTy SCMs reflect real-world mechanisms (or are just inspired by them). Explainability faithfulness is defined with respect to the explained model, and an explanation method should account for the effects of concepts \textit{as they are encoded by the model}, regardless of whether the model learns and represents real causal structures, synthetic structures, or spurious correlations. 

\subsection{Deterministic Decoding}
\label{sub:deterministic}

\textbf{Why is deterministic decoding necessary?}~While deterministic decoding has clear drawbacks, it is essential for LIBERTy. Counterfactual generation requires fixing the exogenous variables of the DGP. Yet, stochastic decoding introduces noise at the token-sampling level that lies outside the DGP and cannot be controlled or recorded. As a result, such counterfactuals cannot serve as `structural' ones. Furthermore, they also fail by intuitive standards. Although the prompt for generating the original example and the counterfactual may differ only in one concept value, stochastic decoding often produces an entirely new narrative with little lexical overlap. While lexical overlap is not formally required, it remains a widely used proxy for counterfactual quality in NLP. Accordingly, many works generate counterfactuals by instructing LLMs to edit the original text minimally \citep{Gat2024, 0002XMZQ24, WangQY0ZF024}. However, such examples are only approximations, since entirely different DGPs produce the original and counterfactual texts. Alternative solutions, beyond our approach of using exogenous grounding texts, include generating multiple counterfactuals and estimating ICaCE by averaging over them, or employing controlled decoding methods for counterfactual generation \citep{ChatziBSTG25, RavfogelSSC25}.

\subsection{LLM-generated Counterfactuals}
\label{sub:counterfactual_fails}

\textbf{Why do explanations based on LLM-generated counterfactuals fail?}~
Explanations based on LLM-generated counterfactuals perform surprisingly well in benchmarks such as CEBaB \citep{Gat2024}, where human annotators provide the reference counterfactuals against which explanations are evaluated. However, this performance stems from the fact that both humans and LLMs approach the task similarly, by minimally editing the input text to reflect a change in concept. In such settings, LLMs can closely mimic the references, particularly when the texts are short and simple. Evaluation using human-written counterfactuals is therefore not an assessment of causal effects, but rather an evaluation of how well models mimic human editing. When evaluated under LIBERTy, however, the limitations of this approach become clear. Unlike human-written counterfactuals, LIBERTy provides structural counterfactuals derived from causal interventions in the DGP. LLM-generated counterfactuals fail in this setting because their edits reflect heuristic assumptions, rather than the actual underlying mechanism. 

\subsection{Opportunities}
\label{sub:opportunities}

\textbf{What are the opportunities in the intersection between causality, explainability, and NLP/LLMs?}~Our findings reveal several exciting opportunities at the intersection of causality, explainability, and NLP. First, we observed a large room for improvement in both local and global explanations, offering clear targets for future work. There is clear potential for the development of causal-inspired explanation methods.
Instead of relying on LLM-based explanations, which, despite encoding broad knowledge in their parameters, are not exposed to data from the target DGP and therefore fail to provide faithful explanations, small but principled techniques offer a more promising direction. These approaches can rely on causal structure rather than scale, making them especially well-suited for academic research.
LIBERTy provides a rigorous evaluation ground for such methods and, we hope, will foster their further development. 

Finally, our analysis revealed that vanilla fine-tuning may fail to capture the causal structure of the data, suggesting the need for unique learning methods. This opens an opportunity to harness LIBERTy as a testbed for developing and benchmarking new causal learning methods that go beyond fine-tuning, approaches that explicitly aim to align models with the underlying DGP. To summarize, there is great promise in developing smaller, theory-grounded, causal-inspired explainability and learning approaches. We hope our work will serve as a foundation for such future research.
\begin{table*}[t]
\centering
\footnotesize
\begin{adjustbox}{width=0.6\textwidth}
\begin{tabular}{lccc|c}
\toprule
\cellcolor{Gray!15} & \cellcolor{Gray!15} & \cellcolor{Gray!15} & \cellcolor{Gray!15} & \cellcolor{Gray!15} \\
\cellcolor{Gray!15} & \multirow{-2}{*}{\cellcolor{Gray!15}\makecell{\textbf{Workplace} \\ \textbf{Violence}}} & \multirow{-2}{*}{\cellcolor{Gray!15}\makecell{\textbf{Disease} \\ \textbf{Detection}}} & \multirow{-2}{*}{\cellcolor{Gray!15}\makecell{\textbf{CV} \\ \textbf{Screening}}} & \multirow{-2}{*}{\cellcolor{Gray!15}\textbf{Avg.}} \\
\midrule
\# Annotators & 6 & 5 & 6 & 5.67 \\
\# Individual &  76 & 170 & 103 & 116.33 \\
\# Pairs &  101 & 105 & 106 & 104 \\
\# Labels & 481 & 955 & 621 & 685.67 \\
\midrule
Avg. IAA &  0.90 & 0.92 & 0.91 & 0.91 \\
Avg. MAE & 0.35 & 0.53 &  0.62 & 0.50 \\
\midrule
Concepts &  97.9\% & 100\% & 84.7\% & 94.2\% \\
Coherence &  4.75 & 4.88 & 4.75 & 4.79 \\
Fluency & 4.72 & 4.90 & 4.92 & 4.85 \\
Relevancy &  4.80 & 4.68 & 4.83 & 4.77 \\
Consistency &  4.92 & 4.92 & 4.92 & 4.92 \\
Plausibility &  4.63 & 4.62 & 4.07 & 4.44 \\
\bottomrule
\end{tabular}
\end{adjustbox}
\caption{\textbf{Results of Human Validation:} Average IAA and MAE are computed across annotator pairs: IAA for the binary concept identification task, and MAE for all other tasks using a 1–5 Likert scale. `Concepts' reports the percentage of concept values that were marked as explicitly stated or logically inferred. `Plausibility' reports the average score for a pair of texts being judged as an original and its counterfactual.}
\label{tab:human_stats}
\end{table*}

\section{Human Validation}
\label{sec:human_val}

We conduct human validation of the generated examples to ensure: (1) they include all concept values; (2) they have high linguistic quality, by measuring coherence and fluency; (3) they are relevant to the task (e.g., look like a personal statement); (4) they are logically consistent with themselves and external facts; (5) the counterfactual feels like a genuine counterfactual (by measuring how likely the text was written by the same person in a parallel world where the concept value is different). Notably, human validation is not required to ensure that the LIBERTy evaluation pipeline is faithful; however, it helps demonstrate that the synthetically generated data is realistic and practical. 

\begin{figure*}[t]
    \centering
    \includegraphics[width=0.98\textwidth]{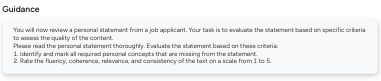}
    \includegraphics[width=0.98\textwidth]{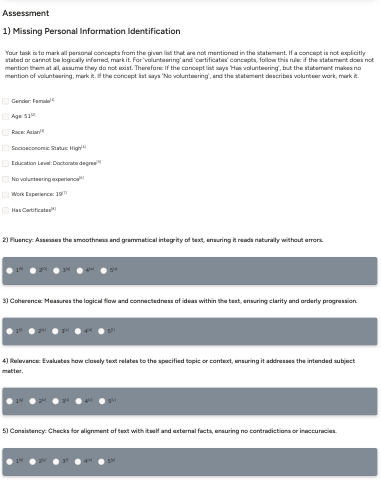}
    \caption{Annotation guidelines for validating concept values and rating coherence, fluency, task relevance, and logical consistency. Example of the CV screening dataset.}
    \label{fig:dim_annotation}
\end{figure*}

\begin{figure*}[t]
    \centering
    \includegraphics[width=0.98\textwidth]{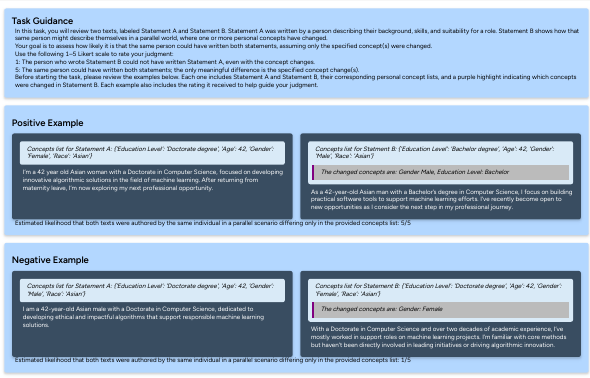}
    \includegraphics[width=0.98\textwidth]{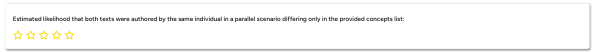}
    \caption{Annotation guidelines for rating the plausibility of a text as a genuine counterfactual of the original.}
    \label{fig:cf_annotation}
\end{figure*}

We recruited 13 annotators (all graduates with fluent English; 3 males, 10 females) who annotated a total of 349 single-text and 312 text-cf-pair evaluations. Each text was rated across six dimensions: five individual attributes assessing text-level quality and one comparative attribute assessing the quality of the counterfactual relative to its original. This resulted in a total of $349 \times 5 + 312 \times 1 = 2{,}057$ labels.
The average inter-annotator agreement (IAA) across all dimensions is 0.91. The annotation guidelines can be viewed in Figures~\ref{fig:dim_annotation} and \ref{fig:cf_annotation}.

The results are presented in Table~\ref{tab:human_stats}. As shown, the generated examples exhibit high linguistic quality, with average scores of 4.79 and 4.85 out of 5 for coherence and fluency, respectively. Their average scores for task relevance and logical consistency are 4.77 and 4.92. In addition, agreement with concept values is 94.2\% on average, indicating that GPT-4o accurately instantiates the sampled values.
The lowest scores appeared in the CV Screening dataset, probably because it involves socially sensitive concepts that are more heavily filtered during generation. Finally, annotators judged the counterfactuals to be genuine, with an average score of 4.44 out of 5, demonstrating that they were perceived as plausible even by the human eye.

\section{Explainability Methods}
\label{sec:explainability}

In this section, we provide additional background on the explainability methods used in our study, as well as further implementation details for each.

\begin{table*}[t]
\centering
\footnotesize
\begin{adjustbox}{width=0.65\textwidth}
\begin{tabular}{l|cc|cc|cc|cc}
\toprule
 \cellcolor{Gray!15}$\rightarrow$ \textbf{Model} &
\multicolumn{2}{c|}{\cellcolor{Gray!15}\textbf{Average}} &
\multicolumn{2}{c|}{\cellcolor{Gray!15}\textbf{DeBERTa-v3 }} &
\multicolumn{2}{c|}{\cellcolor{Gray!15}\textbf{T5
}} &
\multicolumn{2}{c}{\cellcolor{Gray!15}\textbf{Qwen-2.5}} \\
\cellcolor{Gray!15} $\downarrow$ \textbf{Technique} & \cellcolor{Gray!15}ED & \cellcolor{Gray!15}OF & \cellcolor{Gray!15}ED & \cellcolor{Gray!15}OF & \cellcolor{Gray!15}ED & \cellcolor{Gray!15}OF & \cellcolor{Gray!15}ED & \cellcolor{Gray!15}OF \\
\midrule
\textit{Only Change} & 0.59 & 0.49 & 0.54 & 0.51 & 0.50 & 0.50 & 0.72 & 0.46 \\
\textit{Fix All} & \textbf{0.54} & \textbf{0.58} & \textbf{0.46} & \textbf{0.62} & \textbf{0.44} & \textbf{0.61} & 0.72 & \textbf{0.50} \\
\textit{Fix Confounders} & 0.55 & 0.55 & 0.49 & 0.57 & 0.46 & 0.58 & \textbf{0.71} & \textbf{0.50} \\
\textit{Meds \& Confs} & 0.57 & 0.54 & 0.48 & 0.58 & 0.49 & 0.55 & 0.73 & 0.48 \\
\bottomrule
\end{tabular}
\end{adjustbox}
\caption{\textbf{Results of Counterfactual Generation Prompting:} We report the Average Error Distance ($\overline{\mathrm{ED}}$) and Average Order-Faithfulness ($\overline{\mathrm{OF}}$) for the four prompting techniques used in counterfactual generation with Gemini-1.5-Pro. \textit{Meds \& Confs} is Mediators and Confounders: mentioning mediators while instructing to fix confounders.}
\label{tab:cf_results}
\end{table*}

\subsection{Counterfactual Generation}
\label{sub:counterfactuals}

This approach uses an LLM (or a fine-tuned, pre-trained model when parallel training data are available) to generate approximations of counterfactuals. Typically, the LLM is instructed to modify the input text by replacing a specified concept with a target value. \citet{Gat2024} propose injecting causal assumptions into the prompt, in particular identifying confounder concepts from the causal graph and prompting the LLM to keep them fixed while changing the target concept. They found that LLM-generated counterfactuals yielded the best explanation method on CEBaB. In light of this, we extend their approach and compare different prompting strategies, each of which injects distinct causal assumptions into the prompt. In our causal graphs, relative to the target concept being modified, other concepts may play two key roles. The first are confounders, which act as root causes that influence both the target concept and the text, and therefore must remain fixed. The second are mediators, which are influenced by the target concept and, in turn, influence the text. They must be allowed to vary when measuring total causal effects. 

The prompting techniques we evaluate are:
(a) \textit{Only Change}: specifies only the target concept change;
(b) \textit{Fix All}: specifies the change and instructs the LLM to fix the values of all other concepts;
(c) \textit{Fix Confounders}: specifies the change and the causal parents, explicitly forbidding their alteration.;
(d) \textit{Mediators and Confounders}: specifies all mediator concepts (without asking to fix their values) and the change, while instructing the LLM to fix the values of the confounding concepts.

To generate counterfactuals, we use Gemini-1.5-Pro, which differs from the LLM used to generate LIBERTy examples (GPT-4o). Importantly, although the prompts may mention the concepts and sometimes their roles (confounders or mediators), Gemini is expected to infer on its own how a change in the target concept affects other concepts (if they are mediators) and the resulting text. To compare different prompting techniques and manage computational costs, we restrict our experiments to the CV Screening dataset and three fine-tuned models: DeBERTa-base, T5-base, and Qwen2.5-1.5B. The results are reported in Table~\ref{tab:cf_results}. As shown, the best-performing prompting technique is Mediators and Confounders, which is also the most causally informed. This technique explicitly incorporates both causal roles: it asks to hold the confounders fixed while allowing mediators to vary according to Gemini's decision. Since this technique works the best, we use it in all other experiments.
The full set of prompt versions used for this task is provided in Appendix~\ref{sub:prompts-CF-Generation-method}.

\subsection{Matching}
\label{sub:matching}

Although counterfactual generation is a valuable explainability approach, employing LLMs during inference can be costly, either due to latency or financial expenses. An alternative is to use a more efficient method that searches for approximations within a predefined set of candidate texts. This approach, known as matching, involves identifying the most similar candidate text whose target concept corresponds to the desired target value. Matching methods differ in how they perform the search. We evaluate two approaches: matching based on semantic similarity and matching based on concept values. A third approach involves learning causal representations \citep{Gat2024}, which lies outside the scope of our study. In addition, we adopt the top-$k$ matching technique (with $k=3$), which has been shown to outperform single matching \citep{Gat2024}. 

\paragraph{Semantic-based Matching}~For each original text and concept change $C:c \rightarrow c'$, we retrieve the top-$k$ candidates with $C=c'$ based on cosine similarity between mean-pooled text embeddings. To compute embeddings, we examine three encoder-only models: (1) \textit{ST Match}: a SentenceTransformer model (the default `all-MiniLM-L6-v2' model) \citep{ReimersG19}; (2) \textit{PT Match}: a pre-trained DeBERTa model (DeBERTa-base version); and (3) \textit{FT Match}: a DeBERTa model fine-tuned to predict $Y$ in each dataset.

\paragraph{Concept-based Matching}~For each original text and concept change $C:c \rightarrow c'$, we retrieve the top-$k$ candidates with $C=c'$ based on similarity of the remaining concept values. Since we assume that the explanation method does not have direct access to the gold concept labels, we fine-tune a DeBERTa model (DeBERTa-base version) to predict concept values from text. Matching is then performed in two alternative ways: (1) \textit{Approx} — all other concept values must match exactly, with a single mismatch permitted only if no perfect match is available; (2) \textit{ConVecs} — we concatenate the softmax prediction vectors of all concepts into a single vector and compute cosine similarity between this vector for the original example and each candidate.

\subsection{Concept Erasure}
\label{sub:concept_erasure}

Concept erasure methods intervene on a model’s internal representations to remove information about a target concept, typically by projecting out directions in the activation space that encode it. By comparing model behavior before and after erasure, these methods estimate the influence of the concept on predictions. In this study, we evaluate the state-of-the-art erasure method LEACE \citep{belrose2023leace}. LEACE is a closed-form method that removes all linearly encoded information about a target concept, while minimizing distortion to other directions. Given a hidden representation $h(x)$ LEACE computes an affine projection that eliminates the components aligned with the concept direction $v_c$. This yields an erased representation $h^{\text{erased-}c}(x)$. The effect of the concept is then defined as the difference between the model’s predictions for $h(x)$ and on $h^{\text{erased-}c}(x)$.

\textit{Applicability Note:} In our experiments, we apply LEACE by extracting embeddings via mean pooling. Since LEACE assumes that a concept value of 0 corresponds to the concept being absent, we restrict its use to the \textsc{Disease Detection} dataset, where this assumption holds naturally (e.g., symptom absence). In other datasets, the concepts of interest involve changes between two non-null states (e.g., gender, occupation), for which the “absence” assumption does not apply, making erasure ill-defined. Finally, because LEACE requires access to and modification of internal embeddings, we apply it only to fine-tuned models that support this interface: DeBERTa-base, T5-base, and Qwen2.5-1.5B in our evaluation.

\subsection{Concept Attributions}
\label{sub:attributions}

Concept attribution methods map concepts to vectors or subspaces within a model's internal activation space, typically derived from concept-labeled examples. These vectors capture directions in the hidden representation space that the model relies on for prediction, enabling us to quantify how movement along a concept direction affects the model's output and thereby assess the concept's importance. In our experiments, we combine ConceptShap \citep{Human_KeyConcepts} with TCAV \citep{TCAV_Kim2018}, two widely used concept attribution methods in computer vision. ConceptShap quantifies the contribution of concepts to a model's predictive performance using Shapley values. Unlike TCAV, which measures directional sensitivity along a single concept vector, ConceptShap treats concepts as players in a cooperative game and attributes credit to them based on their marginal contributions across all possible coalitions of concepts. To apply this framework, one first requires a representation for each concept and then computes Shapley values. Since our goal is to evaluate predefined concepts, we construct their representations using TCAV vectors. TCAV derives concept vectors by training a linear classifier in the activation space to separate examples that contain the concept from those that do not, and then uses the classifier's normal vector as the concept representation. ConceptShap is then applied over these predefined vectors to assign Shapley-based importance scores.

\textit{Applicability Note:} Both ConceptShap and TCAV are primarily global explanation methods: they quantify how concepts influence the model's predictions across a dataset, rather than for individual inputs. Accordingly, we evaluate them only in the global explainability setup.

\onecolumn

\section{Dataset Details}
\label{sec:dataset_details}

\subsection{Workplace Violence}
\label{sub:workplace_violence}

\subsubsection{SCM}

This dataset simulates HR–nurse interviews, in which the (explained) model predicts the likelihood that a nurse will experience workplace violence. The causal graph is adapted from the Minnesota Nurses' Study \citep{violence_nurse_study}, which documented the prevalence of verbal and physical violence among clinical staff and analyzed risk factors by demographic and professional background. 
We perform minor simplifications to reduce the number of concepts and to rename them for clarity. The simplified version preserves the main causal relations reported in the original paper while maintaining readability.

The template follows a structured HR interview format. To ensure both realism and sufficient diversity, we generate interview templates as follows: for each concept, a bank of 10 questions is created using Gemini, each designed to elicit the concept's value from different linguistic perspectives. Additionally, 10 opening and 10 closing sentence variants are defined to maintain a coherent interview flow. Each template is generated by sampling one question per concept, along with an opening and closing sentence. The question order is randomized, yielding a large pool of interview templates. The persona contains three informal ``fun facts'' about the nurse, each centered on a concept (without specifying its value). Using Gemini, we generated 500 personas.

\begin{table*}[ht]
\renewcommand{\arraystretch}{1.2}
\centering
\normalsize
\begin{adjustbox}{width=0.98\textwidth}
\begin{tabular}{p{0.05\textwidth}p{0.25\textwidth}>{\centering\arraybackslash}m{0.4\textwidth} >{\centering\arraybackslash}m{0.15\textwidth}>{\centering\arraybackslash}m{0.15\textwidth}}
\toprule
\cellcolor{Gray!15} \textbf{$C$} & \cellcolor{Gray!15} \textbf{Name} & \cellcolor{Gray!15} \textbf{Values} & \cellcolor{Gray!15} \textbf{Parents} & \cellcolor{Gray!15} \textbf{Childs} \\
\midrule
$Y$ & Violence Experience & \{0: No Violence, 1: Verbal Violence, 2: Physical Violence\} & all & -- \\
$G$ & Gender & \{0: Female, 1: Male\} & -- & L, D\\
$A$ & Age & \{0: 24--32, 1: 34--44, 2: 46--55\} & -- & T, L \\
$R$ & Race & \{0: African American, 1: Hispanic, 2: White, 3: Asian\} & -- & L, D, S, Y \\
$T$ & Tenur & \{0: 4--9, 1: 10--19, 2: 20--25\} & A & S, Y \\
$L$ & License & \{0: LPN, 1: RN, 2: APRN\} & G, R, A & S, Y \\
$D$ & Department & \{0: Family Practice, 1: ICU, 2: Psychiatric/Mental Health, 3: Emergency\} & G, R & Y \\
$S$ & Seniority & \{0: General Staff, 1: Experienced Staff, 2: Middle Management, 3: Senior Management\} & A, G, R, T, L & Y \\
\midrule
\midrule
\multicolumn{5}{l}{$G \sim \mathrm{Uniform}\{0,1\}$}\\
\multicolumn{5}{l}{$A \sim \mathrm{Categorical}\{0{:}\,25\%,\,1{:}\,50\%,\,2{:}\,25\%\}$}\\
\multicolumn{5}{l}{$R \sim \mathrm{Uniform}\{0,1,2,3\}$}\\
\multicolumn{5}{l}{$T=\min(2,\max(0,\mathrm{round}(0.8\,A+\varepsilon_T))) \qquad \varepsilon_T\sim \mathcal{N}(0.05,\,0.5)$}\\
\multicolumn{5}{l}{$L=\min(2,\max(0,\mathrm{round}(0.3\,G+0.3\,R+0.2\,A+\varepsilon_L))) \qquad \varepsilon_L\sim \mathcal{N}(0,\,0.5)$}\\
\multicolumn{5}{l}{$D=\min(3,\max(0,\mathrm{round}(0.5\,G+0.4\,R+0.4+\varepsilon_D))) \qquad \varepsilon_D\sim \mathcal{N}(0.2,\,0.5)$}\\
\multicolumn{5}{l}{$S=\min(3,\max(0,\mathrm{round}(0.4\,A+0.1\,(G+R)+0.3\,(T+L)+\varepsilon_S))) \qquad \varepsilon_S\sim \mathcal{N}(0,\,0.5)$}\\
\multicolumn{5}{l}{$Y=\min(2,\max(0,\mathrm{round}(0.5\,(G+D)-0.2\,(A+R+L+T+S)+0.8+\varepsilon_Y))) \qquad \varepsilon_Y\sim \mathcal{N}(0.3,\,0.2)$}\\
\bottomrule
\end{tabular}
\end{adjustbox}
\caption{SCM of the Workplace Violence Prediction Dataset.}
\label{tab:scm_workplace}
\end{table*}

\subsubsection{Prompts}
\begin{prompt}[label={box:nurse-persona-prompt}]{LimeGreen}{Nurse Persona Generation Prompt}
\small
\textbf{System Instruction:}\\
\textit{
Your task is to create an engaging nurse persona by generating fun facts for three given aspects. 
These facts should highlight the nurse's professional or personal journey.
}

\textbf{User Prompt:}\\
\textit{
Here are the three aspects: \texttt{\{sample\_aspects[0]\}}, \texttt{\{sample\_aspects[1]\}}, \texttt{\{sample\_aspects[2]\}}.\\
Please creatively generate three surprising and contextually relevant fun facts for each aspect that highlight the nurse's professional or personal journey.\\
Aim to enrich the persona and captivate the audience by revealing unique insights into the nurse’s experiences.\\
\textbf{Respond in this format:}\\
\textbf{Fun Fact on \texttt{\{sample\_aspects[0]\}}}: \\
\textbf{Fun Fact on \texttt{\{sample\_aspects[1]\}}}: \\
\textbf{Fun Fact on \texttt{\{sample\_aspects[2]\}}}: 
}
\end{prompt}
\vspace{-0.5em}

\begin{prompt}[label={box:nurse-cf-generation-prompt}]{LimeGreen}{Original \& Counterfactual Nurse Dialogue Generation Prompt}
\small
\textbf{System Instruction:}\\
\begingroup\itshape
As a specialist in refining dialogues between HR personnel and a nurse, your task is to enhance 
the conversation with added depth, personal insights, and storytelling. 
The primary goal is to remain fully consistent with the nurse's personal information provided. 
You will also be given fun facts about the nurse’s persona. Use these to enrich the dialogue, 
but adjust the facts as needed to ensure they align with the personal information. 
If any fun fact conflicts with the personal information, rewrite it to match. 
Finally, make sure the resulting dialogue feels coherent and natural. 
Avoid repeating questions or asking something that has already been mentioned. 
Ensure that everything flows smoothly, as if it were a real and authentic conversation.
\par\endgroup

\textbf{User Prompt:}\\
\begingroup\itshape
Based on the provided base dialogue, revise the conversation to incorporate more depth and 
include all adjusted fun facts from the nurse's persona. Ensure these fun facts align with the 
nurse's personal information; revise any discrepancies to accurately reflect the nurse's true values.\\[0.5em]

\textbf{Nurse's personal information}: \texttt{\{nurse\_details\}}\\
\textbf{Nurse's Persona}: \texttt{\{nurses\_persona\}}\\
\textbf{Base dialogue}: \texttt{\{dialogue\_draft\}}\\[0.5em]

Final dialogue:
\par\endgroup
\end{prompt}

\subsubsection{Examples}
\begin{prompt}[label={box:nurse-template-example}]{Gray}{Example of Nurse Dialogue Template}
\small
\textit{
\textbf{Intro:} Excited for our chat. I'm from HR, and we've got a brief 5-minute discussion ahead to collect some personal and demographic information. How have you been coping with everything?\\
\textbf{Department Question:} Just for clarity, can you tell us your specific department?\\
\textbf{Department Info:} Intensive Care Unit (ICU)\\
\textbf{Race Question:} How would you describe your race or ethnicity?\\
\textbf{Race Info:} African American\\
\textbf{Age Question:} How old are you, if you're comfortable sharing?\\
\textbf{Age Info:} 44\\
\textbf{Gender Question:} Just to get a clearer picture, could you tell me your gender?\\
\textbf{Gender Info:} Male\\
\textbf{License Type Question:} Could you indicate which nursing license you've obtained? LPN, RN, or APRN?\\
\textbf{License Type Info:} Registered Nurse (RN)\\
\textbf{Years As Nurse Question:} Could you let us know how long you've been working in the nursing field?\\
\textbf{Years As Nurse Info:} 10\\
\textbf{Activity At Work Question:} Could you describe the extent of decision-making authority you hold in your current role?\\
\textbf{Activity At Work Info:} Experienced Staff\\
\textbf{Closing:} Short chat, but it was packed with insights. Thanks for helping me out. Bye for now.
}
\end{prompt}
\begin{prompt}[label={box:nurse-persona-example}]{Gray}{Example of Generated Nurse Persona}
\small
\textit{
\textbf{Fun Fact on Race:} While studying to become a nurse, she volunteered at a free clinic that primarily served a community different from her own, opening her eyes to healthcare disparities and solidifying her commitment to equitable care.\\
\textbf{Fun Fact on Age:} She decided to become a nurse later in life, after a successful career in a completely unrelated field, bringing a unique perspective and wealth of life experience to her nursing practice.\\
\textbf{Fun Fact on Department:} Although she always envisioned herself working in a fast-paced environment like the ER, she discovered a passion for palliative care, finding deep fulfillment in providing comfort and support to patients and their families during challenging times.
}
\end{prompt}
\begin{prompt}[label={box:nurse-cf-example}]{Gray}{Example of Original and Counterfactual Generated Nurse Dialogue}
\small

\textit{\textbf{Changed aspect:} Gender from Male to Female and Department from ``Psychiatric/Mental Health'' to ``Intensive Care Unit (ICU)''.}\\

\textbf{Original Dialogue:}\\
\textbf{HR:} Hi James, I'm excited for our chat today. I'm from HR, and we've got a brief 5-minute discussion ahead to collect some personal and demographic information. How have you been coping with everything?\\
\textbf{James:} Hi! I've been doing well, thank you. It's been a busy time, but I'm managing to keep everything balanced.\\
\textbf{HR:} That's good to hear. Let's dive in. Could you tell me about your department? What challenges and rewards come with working there?\\
\textbf{James:} I work in the Psychiatric/Mental Health department. It's a challenging field, but incredibly rewarding. Every day, I get to make a difference in people's lives, helping them navigate their mental health journeys. It's a privilege to support them through their toughest times.\\
\textbf{HR:} That sounds like meaningful work. Just to get a clearer picture, could you tell me your gender?\\
\textbf{James:} Sure, I'm male.\\
\textbf{HR:} And would it be okay to tell me your age?\\
\textbf{James:} I'm 43 years old. Turning 40 was a milestone for me, and it inspired me to fulfill a lifelong dream of learning Spanish. It's been incredibly rewarding to connect with my Hispanic patients on a deeper level.\\
\textbf{HR:} That's wonderful! How would you describe your race or ethnicity?\\
\textbf{James:} I'm African American. My heritage is an important part of who I am, and I strive to bring that perspective into my work every day.\\
\textbf{HR:} Thank you for sharing that. I'm interested in knowing, how many years have you been nursing?\\
\textbf{James:} I've been a nurse for 10 years now. It's been a fulfilling journey, and I continue to learn and grow every day.\\
\textbf{HR:} Could you tell us whether you are an LPN, RN, or APRN?\\
\textbf{James:} I'm a Licensed Practical Nurse (LPN). I initially trained as an RN, but I found my calling as an LPN, where I can have a direct impact on patient care.\\
\textbf{HR:} Could you describe the extent of decision-making authority you hold in your current role?\\
\textbf{James:} As an experienced staff member, I have a significant role in decision-making, especially when it comes to patient care plans. I work closely with the team to ensure we're providing the best possible support for our patients.\\
\textbf{HR:} It sounds like you have a lot of responsibility and passion for your work. Short chat, but it was packed with insights. Thanks for helping me out. Bye for now.\\
\textbf{James:} Thank you! It was great talking to you.\\[1em]

\textbf{CF Dialogue:}\\
\textbf{HR:} Hi, I'm excited for our chat today. I'm from HR, and we've got a brief 5-minute discussion ahead to collect some personal and demographic information. How have you been coping with everything?\\
\textbf{Nurse:} Hi there! I've been managing well, thank you. The ICU can be intense, but it's also incredibly rewarding.\\
\textbf{HR:} I can imagine. Tell me about your department. What challenges and rewards come with working there?\\
\textbf{Nurse:} I work in the Intensive Care Unit (ICU). It's a challenging environment because we deal with critically ill patients, but it's rewarding to see them recover and know that I played a part in their journey. The teamwork here is phenomenal, and we support each other through the toughest days.\\
\textbf{HR:} That sounds like a very fulfilling role. Just to get a clearer picture, could you tell me your gender?\\
\textbf{Nurse:} Sure, I'm female.\\
\textbf{HR:} Would it be okay to tell me your age?\\
\textbf{Nurse:} I'm 43 years old. It's been an interesting journey, especially since turning 40, which inspired me to start learning Spanish. It's been wonderful to connect with my Hispanic patients on a deeper level.\\
\textbf{HR:} That's impressive! How many years have you been nursing?\\
\textbf{Nurse:} I've been a nurse for 10 years now. I started my career as an LPN, and it's been a fulfilling decade of learning and growth.\\
\textbf{HR:} How would you describe your race or ethnicity?\\
\textbf{Nurse:} I'm African American. My heritage is an important part of who I am, and I strive to bring that perspective into my work every day.\\
\textbf{HR:} Could you describe the extent of decision-making authority you hold in your current role?\\
\textbf{Nurse:} As an experienced staff member, I have a significant amount of decision-making authority. I often collaborate with doctors and other nurses to determine the best care plans for our patients. It's a role that requires both leadership and teamwork.\\
\textbf{HR:} Could you tell us whether you are an LPN, RN, or APRN?\\
\textbf{Nurse:} I'm a Licensed Practical Nurse (LPN). I initially trained as an LPN because I wanted to get into the field quickly and start making a difference. It's been a rewarding path, and I continue to learn every day.\\
\textbf{HR:} Short chat, but it was packed with insights. Thanks for helping me out. Bye for now.\\
\textbf{Nurse:} Thank you! It was great talking to you. Have a wonderful day!\\

\end{prompt}

\subsection{Disease Detection}

\label{sub:disease_detection}

\subsubsection{SCM}

This dataset simulates clinical self-reports, where the (explained) model predicts a disease from symptoms described in a medical forum post. Unlike the other two datasets, the learning problem is anti-causal: the disease label serves as the root cause in the SCM and determines the values of symptom concepts, based on known symptom–disease relations \citep{monto2000clinical, cady2002sinus}. We also used we used domain knowledge from the Cleveland Clinic\footnote{\url{https://my.clevelandclinic.org/health/diseases}} to identify the key symptoms associated with each condition. Each disease node serves as a parent node to its characteristic symptoms, some of which overlap across diseases to introduce realistic confounding. Dependencies between symptoms (e.g., bright light affecting headache) were explicitly modeled as causal edges. Additionally, symptom prevalence was modeled in the SCM functions, such that more characteristic symptoms have stronger causal weights (e.g., facial pain is more likely than fever for sinusitis).

The template is a narrative structure abstracted from 1,310 posts on Reddit's DiagnoseMe forum,\footnote{https://www.reddit.com/r/DiagnoseMe/} using Gemini to preserve the clinical tone and flow. The persona (a total of 1200) consists of three informal facts about occupation, hobbies, and family or friends. To generate personas, we first sample an occupation and a hobby from predefined lists, then use Gemini to generate the corresponding facts. Each dataset example is created by prompting GPT-4o to follow the template and integrate information from the persona and the symptom values.

\begin{table*}[!ht]
\renewcommand{\arraystretch}{1.2}
\centering
\normalsize
\begin{adjustbox}{width=0.98\textwidth}
\begin{tabular}{p{0.05\textwidth}p{0.25\textwidth}>{\centering\arraybackslash}m{0.4\textwidth} >{\centering\arraybackslash}m{0.15\textwidth}>{\centering\arraybackslash}m{0.15\textwidth}}
\toprule
\cellcolor{Gray!15} \textbf{$C$} & \cellcolor{Gray!15} \textbf{Name} & \cellcolor{Gray!15} \textbf{Values} & \cellcolor{Gray!15} \textbf{Parents} & \cellcolor{Gray!15} \textbf{Childs} \\
\midrule
$Y$ & Disease & \{0: Migraine, 1: Sinusitis, 2: Influenza\} & -- & all \\
$D$ & Dizziness & \{0: Absent, 1: Mild, 2: Strong\} & Y & -- \\
$L$ & Light Sensitivity & \{0: Absent, 1: Mild, 2: Strong\} & Y & H \\
$P$ & Facial Pain & \{0: Absent, 1: Mild, 2: Strong\} & Y & -- \\
$W$ & Weakness & \{0: Absent, 1: Mild, 2: Strong\} & Y & -- \\
$F$ & Fever & \{0: Absent, 1: Mild, 2: Strong\} & Y & -- \\
$N$ & Nasal Congestion & \{0: Absent, 1: Mild, 2: Strong\} & Y & H \\
$H$ & Headache & \{0: Absent, 1: Mild, 2: Strong\} & Y, L, N & -- \\
\midrule
\midrule
\multicolumn{5}{l}{$Y=\varepsilon_Y,\quad \varepsilon_Y \sim \mathrm{Cat}(\{0:\tfrac{1}{3},\,1:\tfrac{1}{3},\,2:\tfrac{1}{3}\})$} \\
\multicolumn{5}{l}{$D=\min(2,\max(0,\mathrm{round}(0.9 \cdot \mathbf{1}\{Y=0\} + \varepsilon_D))),\quad \varepsilon_D\sim \mathcal{N}(-0.1,\,0.6)$} \\
\multicolumn{5}{l}{$L=\min(2,\max(0,\mathrm{round}(0.9 \cdot \mathbf{1}\{Y=0\} + \varepsilon_L))),\quad \varepsilon_L\sim \mathcal{N}(0.2,\,0.5)$} \\
\multicolumn{5}{l}{$N=\min(2,\max(0,\mathrm{round}(0.7 \cdot \mathbf{1}\{Y=1\} + 0.4 \cdot \mathbf{1}\{Y=2\} + \varepsilon_N))),\quad \varepsilon_N\sim \mathcal{N}(0,\,0.7)$} \\
\multicolumn{5}{l}{$P=\min(2,\max(0,\mathrm{round}(0.8 \cdot \mathbf{1}\{Y=1\} + \varepsilon_P))),\quad \varepsilon_P\sim \mathcal{N}(0.2,\,0.6)$} \\
\multicolumn{5}{l}{$F=\min(2,\max(0,\mathrm{round}(0.4 \cdot \mathbf{1}\{Y=1\} + 0.6 \cdot \mathbf{1}\{Y=2\} + \varepsilon_F))),\quad \varepsilon_F\sim \mathcal{N}(0,\,0.6)$} \\
\multicolumn{5}{l}{$W=\min(2,\max(0,\mathrm{round}(0.7 \cdot \mathbf{1}\{Y=2\} + \varepsilon_W))),\quad \varepsilon_W\sim \mathcal{N}(0.2,\,0.6)$} \\
\multicolumn{5}{l}{$H=\min(2,\max(0,\mathrm{round}(0.7 \cdot \mathbf{1}\{Y=0\} + 0.4 \cdot \mathbf{1}\{Y=1\} + 0.3L + 0.3N + \varepsilon_H))),\quad \varepsilon_H\sim \mathcal{N}(-0.1,\,0.5)$} \\
\bottomrule
\end{tabular}
\end{adjustbox}
\caption{SCM of the Disease Detection Dataset.}
\label{tab:scm_disease}
\end{table*}

\subsubsection{Prompts}
\begin{prompt}[label={box:disease-template-generation}]{LimeGreen}{Disease Template Generation Prompt}
\small
\textbf{System Instruction:}\\
\textit{
"Develop a narrative template based on the structure of the provided example. The template should abstract the formatting and key transitions from the example, while seamlessly integrating occupation and hobby details into the narrative. Use this template to ensure that any future persona creation maintains the coherence and style of the original example, yet allows for flexibility to adapt to different personas and symptoms."
}\\[0.5em]
\textbf{User Prompt:}\\
\textit{
\textbf{**Analyze Example Format**:} \texttt{\{reddit\_comment\}}\\
From the example provided, analyze and extract the fundamental structure and style used in composing the narrative:\\
1. \textbf{Analyze Example Format:} Focus on how the example is constructed, noting key phrases, transitions, the arrangement of topics, and how personal details are woven into the narrative.\\
2. \textbf{Craft a Template:} Using your analysis, create a narrative template that includes placeholders or cues for integrating occupation and hobby. Ensure the template can be easily adapted to different scenarios while maintaining the style and coherence of the example.\\
\textbf{Your Task:} Generate a narrative template that can be used to create engaging and coherent personas based on any set of personal details, following the style and structure of the example provided.}
\end{prompt}
\begin{prompt}[label={box:disease-persona-prompt}]{LimeGreen}{Disease Persona Generation Prompt}
\small
\textbf{System Instruction:}\\
\textit{
"Your task is to create an engaging persona by generating three interesting facts 
covering their occupation, hobby, and personal life, based on the provided hobby and disease context."
}\\[0.5em]
\textbf{User Prompt:}\\
\textit{
Create an engaging persona using the provided details:\\
\textbf{Persona's occupation:} \texttt{\{occupation\}}\\
\textbf{Persona's hobby:} \texttt{\{hobby\}}\\
\textbf{** Respond in this format **:}\\
\textbf{Occupation:} Detail the persona's job and an interesting related fact/story. 1-2 sentences. \\
\textbf{Hobby:} Describe the persona's hobby and how it enriches their life. 1-2 sentences. \\
\textbf{Family/Friends:} Share a brief story or fact about the persona's interactions with family or friends. 1-2 sentences.
}
\end{prompt}
\begin{prompt}[label={box:disease-cf-generation-prompt}]{LimeGreen}{Original \& Counterfactual Disease Text Generation Prompt}
\small
\textbf{System Prompt:}\\
\textit{
You are an AI assistant tasked with crafting a detailed consultation post for a patient seeking online medical advice.
The consultation should be developed by integrating the patient's provided symptoms, tailored persona details,
and the structural guidance provided by the narrative template. It is essential to explicitly incorporate each symptom
and aspect of the patient's personal background into the post. Your goal is to create a ready-to-submit, engaging,
and clear consultation request that effectively and compellingly explains the patient's situation.
}\\[0.5em]

\textbf{User Prompt:}\\
\textit{
Compose an engaging and detailed consultation post using the following elements:\\
1. \textbf{Narrative Template}: Use the provided template as a guiding framework to structure your consultation. It should shape the flow and organization of the post, ensuring a logical presentation of your symptoms and background story.\\
2. \textbf{Patient's Symptoms List}: This is the most crucial component—it includes the patient's symptoms, which should be described in detail, focusing on their impact on daily activities and overall well-being.\\
3. \textbf{Persona Details}: Enhance the narrative by incorporating persona details, such as lifestyle, hobbies, and family context, to give depth to the post. Explain how the symptoms affect specific aspects of the patient's life.\\
\textbf{Narrative Template}: \texttt{\{reddit\_template\}}\\
\textbf{Patient's Symptoms List}: \texttt{\{verbal\_symptoms\_list\}}\\
\textbf{Persona Details}: \texttt{\{persona\_info\}}\\[0.5em]
Please ensure that the final output is a cohesive and engaging narrative without distinct section breaks.\\
It should be medically informative and follow a logical flow, starting with an introduction that captures the reader's attention,\\
clearly explaining the symptoms and their impact, and concluding with a request for advice or further action.
}
\end{prompt}

\subsubsection{Examples}
\begin{prompt}[label={box:disease-template-example}]{Gray}{ Example of Generated Disease Narrative Template}
\small
\begingroup\itshape
\textbf{Narrative Template for Persona Creation:}\par\vspace{0.25em}

\textbf{1. Opening Statement (Expressing Frustration \& Seeking Help):}\par
``I know this might be a lot, but [briefly explain the challenge of summarizing your symptoms, e.g., they feel scattered, doctors haven't found a solution yet]. It's been incredibly difficult to figure out where to even begin, and I'm feeling incredibly [emotion, e.g., overwhelmed, hopeless, lost]. The doctors I've seen have mainly focused on treating individual symptoms without getting to the root of the problem. I'm desperate for answers and wondering if there are any tests or specialists you could recommend.''\par\vspace{0.5em}

\textbf{2. Known Medical History (Concise \& Factual):}\par
\textbf{Existing Conditions:} [List diagnosed conditions, including year of diagnosis if relevant].\par
\textbf{Current Medications:} [List medications, dosage, and what they are taken for].\par\vspace{0.5em}

\textbf{3. Lifestyle (Brief \& Relevant):}\par
Briefly describe lifestyle factors that could be relevant to health, e.g., smoking, alcohol consumption, diet].
\par\endgroup
\end{prompt}
\begin{prompt}[label={box:disease-persona-example}]{Gray}{Example of Generated Disease Persona}
\small
\begingroup\itshape
\textbf{Occupation:} As an Occupational Health and Safety Technician, they ensure workplaces are safe for everyone. They once investigated a case where someone nearly got stuck in a tunnel, highlighting the importance of their job.\par\vspace{0.5em}

\textbf{Hobby:} Building tunnels as a hobby lets them apply their professional knowledge in a fun, challenging way. Plus, it's incredibly satisfying to create underground spaces.\par\vspace{0.5em}

\textbf{Family/Friends:} Their friends often joke about needing hard hats and safety briefings before visiting, but secretly, they're fascinated by their hobby.
\par\endgroup
\end{prompt}
\begin{prompt}[label={box:disease-cf-example}]{Gray}{Example of Original and Counterfactual Disease text}
\small
\textit{\textbf{Changed aspect:} \relax{Remove: General\_Weakness (Slight)}}\\

\textbf{Original disease's symptom list:} {\relax [Dizzy (Slight), Sensitivity\_to\_Light (Slight), Headache (Slight), Fever (Strong), General\_Weakness (Slight)]}\\

\textbf{Original disease text:}\\
Hi, I'm a 34-year-old Caucasian female living in Portland, Oregon. Lately, I've been struggling with feeling slightly off-balance, and it's been really difficult to maintain my usual pace at work and enjoy my hobbies. This all started about three weeks ago, although thinking back, it might have been creeping up on me for a while. As a psychiatric technician, my role involves providing compassionate care and support to individuals facing mental health challenges. However, this minor instability, especially when moving quickly, has made it challenging to perform my duties with the same confidence and energy. Additionally, I've been feeling slightly fatigued, more tired than usual, which further complicates my ability to be fully present and engaged with my patients.

The symptoms don't stop there. I've noticed a strain in bright light, a subtle discomfort in my eyes when I'm in well-lit areas, which makes it hard to focus during meetings or when documenting patient progress. On top of that, I've been experiencing intense feverishness, feeling extremely warm and very ill at times, which is quite alarming. A mild headache, a gentle but persistent discomfort, accompanies these symptoms, making it difficult to concentrate on tasks or enjoy my favorite pastime, birdwatching. Birdwatching has always been my peaceful escape, where I find solace in identifying different species and often rise before dawn to catch a glimpse of rare birds. However, these symptoms have made it hard to enjoy these early morning adventures, especially when I share them with my niece and nephew, who I love introducing to the wonders of nature.

I'm increasingly concerned about how these symptoms are impacting my daily life and my ability to create special memories with my family. Any advice would be greatly appreciated. Please assist me in returning to the vibrant life I cherish—filled with meaningful work, the joy of birdwatching, and precious moments with my loved ones.\\

\textbf{Counterfactual disease's symptom list:} {\relax [Dizzy (Slight), Sensitivity\_to\_Light (Slight), Headache (Slight), Fever (Strong)]}\\

\textbf{Counterfactual disease text:}\\
Hi, I'm a 34-year-old Caucasian female living in Portland, Oregon. Lately, I've been struggling with feeling slightly off-balance, and it's been really difficult to maintain my usual routine, especially at work and during my cherished hobbies. This all started about three weeks ago, although thinking back, it might have been creeping up on me for a while.

As a psychiatric technician, my role involves providing compassionate care and support to individuals facing mental health challenges. I find deep fulfillment in helping patients develop coping mechanisms and life skills. However, this minor instability, especially when moving quickly, has made it challenging to perform my duties effectively. I often feel a subtle discomfort in my eyes in well-lit areas, which adds to the strain during my shifts. The intense feverishness I experience makes me feel extremely warm and very ill, further complicating my ability to focus and be present for my patients. Additionally, a mild headache lingers, a gentle but persistent discomfort that seems to accompany me throughout the day.

Outside of work, birdwatching has always been my peaceful escape. I love rising before dawn to catch a glimpse of rare birds, finding solace in identifying different species. However, the slight off-balance feeling and the strain in bright light have made these early morning excursions less enjoyable and more challenging. I also cherish sharing this passion with my niece and nephew, creating special memories on nature walks and fostering a love for the natural world. Yet, the symptoms have made it difficult to keep up with their youthful energy and enthusiasm.

I'm reaching out for advice because these symptoms are increasingly impacting my daily life and the activities I hold dear. Any guidance or suggestions would be greatly appreciated. Please assist me in returning to the vibrant life I cherish—filled with meaningful work, peaceful birdwatching, and joyful moments with my family.
\end{prompt}

\subsection{CV Screening}
\label{sub:cv_screening}

\subsubsection{SCM}

This dataset simulates automated resume assessment, where the model is tasked with predicting an applicant's quality from a CV-style personal statement, with labels such as weak, qualified, and outstanding. Motivated by critiques of real-world screening systems \citep{dastin2018amazon, raghavan2020mitigating, cowgill2021biased}, the causal graph encodes hypothesized dependencies between demographic and professional attributes, inspired by statistical patterns reported by the U.S. Bureau of Labor Statistics.\footnote{\url{https://www.bls.gov/cps/demographics.htm}} For example, gender influences the hiring label only indirectly through mediators such as education and Work Experience. We examined multiple demographic and behavioral graphs to infer general causal tendencies, such as differences in education continuation or volunteering rates across demographic groups. 

1,235 templates were generated from 342 scraped personal statement examples,\footnote{\url{https://universitycompare.com}}
where each source text was abstracted with Gemini using a 2-shot prompt to produce several occupation-agnostic variants that preserve the narrative structure while removing concept- and role-specific details.
To generate a persona (a total of 990), we sample a role from a predefined list and use Gemini with a 2-shot prompt to produce both personal and professional context, including motivations and skills relevant to that role. Each dataset example is then created by prompting GPT-4o to follow the template and integrate information from the application role, the persona, and the sampled concept values. 

\begin{table*}[ht]
\renewcommand{\arraystretch}{1.2}
\centering
\normalsize
\begin{adjustbox}{width=0.98\textwidth}
\begin{tabular}{p{0.05\textwidth}p{0.25\textwidth}>{\centering\arraybackslash}m{0.4\textwidth} >{\centering\arraybackslash}m{0.15\textwidth}>{\centering\arraybackslash}m{0.15\textwidth}}
\toprule
\cellcolor{Gray!15} \textbf{$C$} & \cellcolor{Gray!15} \textbf{Name} & \cellcolor{Gray!15} \textbf{Values} & \cellcolor{Gray!15} \textbf{Parents} & \cellcolor{Gray!15} \textbf{Childs} \\
\midrule
$G$ & Gender & \{0: Female, 1: Male\} & -- & E \\
$R$ & Race & \{0: Black, 1: Hispanic, 2: White, 3: Asian\} & -- & E \\
$A$ & Age Group & \{0: 24--32, 1: 33--44, 2: 45--55\} & -- & E, S, W \\
$E$ & Education & \{0: High School, 1: Bachelor’s, 2: Master’s, 3: Doctorate\} & G, R, A & S, W, V, C, Q \\
$S$ & Socioeconomic Status & \{0: Low, 1: Medium, 2: High\} & E, A & V \\
$W$ & Work experience & \{0: 2--5 yrs, 1: 6--10 yrs, 2: 11--25 yrs\} & A, E & C, Q \\
$V$ & Volunteering & \{0: No, 1: Yes\} & E, S & Q \\
$C$ & Certificates & \{0: No, 1: Yes\} & E, W & Q \\
$Q$ & Quality & \{0: Not recommended, 1: Potential hire, 2: Recommended\} & E, V, C, W & -- \\
\midrule
\midrule
\multicolumn{5}{l}{$R=\varepsilon_R \quad \varepsilon_R\sim \text{Uniform}\{0,1,2,3\}$} \\
\multicolumn{5}{l}{$G=\varepsilon_G \quad \varepsilon_G\sim \text{Uniform}\{0,1\}$} \\
\multicolumn{5}{l}{$A=\varepsilon_A \quad \varepsilon_A\sim \text{Categorical}\{0:0.25,\;1:0.50,\;2:0.25\}$} \\
\multicolumn{5}{l}{$E=\min(3,\max(0,\text{round}(0.4\cdot(R{+}A{+}G)+\varepsilon_E))) \quad \varepsilon_E\sim \mathcal{N}(0.35,0.5)$} \\
\multicolumn{5}{l}{$S=\min(2,\max(0,\text{round}(0.45\cdot E+0.25\cdot A+\varepsilon_S))) \quad \varepsilon_S\sim \mathcal{N}(0.25,0.35)$} \\
\multicolumn{5}{l}{$W=\min(2,\max(0,\text{round}(0.5\cdot A+0.3\cdot E+\varepsilon_W))) \quad \varepsilon_W\sim \mathcal{N}(0,0.5)$} \\
\multicolumn{5}{l}{$V=\min(1,\max(0,\text{round}(0.2\cdot E+0.3\cdot S+\varepsilon_V))) \quad \varepsilon_V\sim \mathcal{N}(-0.35,0.2)$} \\
\multicolumn{5}{l}{$C=\min(1,\max(0,\text{round}(0.15\cdot(E+W)+\varepsilon_C))) \quad \varepsilon_C\sim \mathcal{N}(0,0.3)$} \\
\multicolumn{5}{l}{$Q=\min(2,\max(0,\text{round}(0.3\cdot(E+V+C+W)+\varepsilon_Q))) \quad \varepsilon_Q\sim \mathcal{N}(0,0.3)$} \\
\bottomrule
\end{tabular}
\end{adjustbox}
\caption{SCM of CV Screening Dataset.}
\label{tab:scm_cv_screening}
\end{table*}

\subsubsection{Prompts}
\begin{prompt}[label={box:cv-template-generation}]{LimeGreen}{CV Template Generation Prompt}
\small
\textbf{System Instruction:}\\
\textit{
Create a short CV narrative template from the given personal statement example, 
distilling its essential structure and style. 
The template should include key transitions and be concise yet comprehensive, 
ensuring it can adapt to a variety of professional and personal profiles 
while preserving coherence and flexibility.
}\\[0.5em]
\textbf{User Prompt:}\\
\textit{
\textbf{Analyze Personal Statement}: \texttt{sampled\_statement} \\
From the personal statement provided, analyze and extract the fundamental structure and style: \\
1. \textbf{Structure Analysis}: Note key phrases, transitions, and arrangement of professional and personal information. \\
2. \textbf{Template Development}: Using your analysis, create a narrative template weaving qualifications and achievements into a cohesive story. \\
Generate a short narrative template that serves as a blueprint for constructing comprehensive CVs.
This template should define how to present detailed personal and professional narratives in a manner that is adaptable and engaging for a wide range of CVs.
}
\end{prompt}

\begin{prompt}[label={box:cv-persona-prompt}]{LimeGreen}{CV Persona Generation Prompt}
\small
\textbf{System Instruction:}\\
\textit{
Develop a captivating CV persona. Create three compelling facts that weave together personal and professional details, enhancing a CV's appeal. 
Focus on the persona's career motivation, a standout professional ability, and an engaging anecdote linking their family to their career.
}\\[0.5em]
\textbf{User Prompt:}\\
\textit{
Create an engaging persona for the job title '\{job\_title\}'.\\
\textbf{Respond in this format}:\\
\textbf{Motivation for Career Choice}: [Explain what inspired the persona to pursue this career path, linking personal passions with professional goals. 1--2 sentences.]\\
\textbf{Defining Professional Skill}: [Identify a key skill or expertise that highlights the persona's professional capabilities and how it benefits their role. 1--2 sentences.]\\
\textbf{Family and Job Connection}: [Share a memorable moment involving the persona's family that occurred during work, a work-related vacation, or through a work connection. This could include funny incidents, serendipitous meetings of family members via work contexts, or shared experiences directly related to the persona's job. 1--2 sentences.]\\
Ensure that these details are crafted to be adaptable across various demographic and professional attributes, providing a CV that is engaging and rich in content.
}
\end{prompt}

\begin{prompt}[label={box:cf-generation-prompt}]{LimeGreen}{Original \& Counterfactual CV Generation Prompt}
\small
\textbf{System Instruction:}\\
\textit{
You are an AI assistant tasked with crafting a CV Personal Statement for a specific candidate's job application. 
This statement should be developed by integrating the candidate's actual personal information, 
tailored persona details that align with the job role, 
and the structural guidance provided by the narrative template. 
It is essential to explicitly incorporate each piece of the candidate's personal information into the statement. 
The final document should be a ready-to-submit, fluent Personal Statement that is clear, 
aligned with the job level, and effectively conveys the candidate’s suitability for the position through a compelling personal narrative.
}\\[0.5em]
\textbf{User Prompt:}\\
\textit{
Create an engaging CV Personal Statement for a job application using the following elements: \\
1. \textbf{Narrative Template}: Use the provided template as an internal guide. It should influence the flow and organization of the narrative without dictating the final format.\\
2. \textbf{Candidate's Personal Information}: This is the most crucial component. Ensure that every piece of this information is explicitly mentioned and seamlessly woven into the statement. Adjust persona or template details if needed for coherence.\\
3. \textbf{Persona Details}: Enhance the narrative by incorporating persona details, including career choices, required skills, and personal connections to the profession.\\
\textbf{Narrative Template}: \texttt{\{cv\_template\}}\\
\textbf{Candidate's Personal Information}: \texttt{\{candidate\_info\}}\\
\textbf{Persona Details}: \texttt{\{persona\_details\}}\\
Please ensure the final output is a fully-prepared Personal Statement that is fluent and engaging. 
It should start in a unique and captivating manner (avoid beginning with ``from'' or ``as''), 
form a cohesive text that integrates all specified details, 
adhere to the appropriate language style for the job level, 
and present a unified narrative capturing the candidate's story.
}
\end{prompt}

\subsubsection{Examples}

\begin{prompt}[label={box:cv-template-example}]{Gray}{Example of Generated CV Template}
\small
\textit{
Key Points:\\
\textbf{Opening Hook:} Starts with a powerful quote to introduce the overarching interest in psychology.\\
\textbf{Motivating Experience:} Uses a personal experience (Auschwitz trip) to highlight a specific area of interest within Psychology (human behavior).\\
\textbf{Academic Journey:} Chronologically details relevant academic experiences, linking them back to the main interest.\\
\textbf{Skill Demonstration:} Presents extracurricular activities and volunteering experiences to illustrate key skills like communication, teamwork, and problem-solving.\\
\textbf{Real-World Application:} Shares insights from work experience, connecting them to academic knowledge and further solidifying career aspirations.\\
\textbf{Passion Projects:} Highlights personal interests and hobbies, demonstrating well-roundedness and a commitment to personal development.\\
\textbf{Closing Statement:} Reiterates the core motivation and emphasizes personal qualities that make the applicant suitable for the chosen field.\\
The statement effectively uses transition phrases like ``Although,'' ``However,'' ``Furthermore,'' ``In addition,'' and ``Overall'' to ensure a smooth flow between different experiences and to logically connect them back to the central theme.
}
\end{prompt}

\begin{prompt}[label={box:cv-persona-example}]{Gray}{Example of Generated CV Persona}
\small
\textit{
\textbf{Job Title:} Biotech Equity Research Associate\\
\textbf{Motivation for Career Choice:} Driven by a lifelong fascination with the elegance of biological systems and a passion for financial markets, I'm drawn to a career that bridges scientific innovation with sound investment strategies.\\
\textbf{Defining Professional Skill:} My strength lies in distilling complex scientific data into clear, concise, and actionable financial insights, allowing me to effectively communicate opportunities and risks to stakeholders.\\
\textbf{Family and Job Connection:} While attending a biotech conference with my family, we had a delightful encounter with a scientist whose research I had been following. Discussing their groundbreaking work with them (and my starstruck family) solidified my passion for this industry.
}
\end{prompt}
\begin{prompt}[label={box:cf-example}]{Gray}{Example of Original and Counterfactual CV}
\small
\textit{\textbf{Changed concepts:} \relax{Socioeconomic Status from Low to Medium}}\\

\begingroup\itshape
\textbf{Original CV concepts' list:} {\relax [Race: Black/African American, Relevant Work Experience: 9, Gender: Female, Education Level: High School, Socioeconomic Status: Low, Age: 41]}\\

\textbf{Original Resume:}\\
In the dynamic realm of digital marketing, where creativity meets analytics, I have found my professional calling. As a 41-year-old Black/African American woman with nine years of experience in paid media, I am driven by a passion for crafting campaigns that not only captivate audiences but also deliver measurable results. My journey in this field is fueled by a commitment to leveraging data-driven insights to optimize every aspect of a campaign, ensuring maximum return on investment.\\
My career began shortly after high school, where I quickly discovered my knack for analyzing data and translating it into actionable marketing strategies. Over the years, I have honed my skills in data-driven decision-making, becoming adept at analyzing campaign metrics and swiftly pivoting strategies to meet evolving market demands. This ability has been instrumental in my success, allowing me to consistently deliver high-impact results for my clients.\\
One of my defining professional skills is my capacity to balance analytical rigor with creative innovation. This unique combination has enabled me to develop and implement A/B testing strategies that have significantly enhanced campaign performance. My family often jokes about the time I paused a family dinner to excitedly explain a new testing strategy I was implementing, a testament to my deep-seated enthusiasm for my work.\\
Despite coming from a low socioeconomic background, I have persevered and thrived in the fast-paced world of digital marketing. My journey has been marked by a relentless pursuit of excellence and a desire to make a meaningful impact in the industry. I am eager to bring my expertise and passion to your team, contributing to innovative marketing solutions that drive success and growth.\\

\textbf{Original CV concepts' list:} {\relax [Race: Black/African American, Relevant Work Experience: 9, Gender: Female, Education Level: High School, Socioeconomic Status: Medium, Age: 41]}\\

\textbf{Counterfactual Resume:}\\
In the dynamic realm of digital marketing, where creativity meets analytics, I have found my professional calling. As a 41-year-old Black/African American woman with nine years of experience in the field, I am driven by a passion for paid media and a commitment to delivering measurable results. My journey in digital marketing has been fueled by an innate ability to blend data-driven decision-making with creative strategy, ensuring every campaign I manage achieves optimal return on investment.\\
My career began shortly after high school, where I quickly discovered my knack for analyzing campaign metrics and adapting strategies to maximize impact. Over the past nine years, I have honed this skill, becoming adept at swiftly pivoting strategies based on real-time data insights. This ability has not only enhanced my professional growth but has also led to significant achievements, such as increasing client engagement and boosting brand visibility across various platforms.\\
Beyond the numbers, my work is deeply personal. My family often jokes about the time I paused a family dinner to share my excitement over a new A/B testing strategy I was implementing. This anecdote perfectly encapsulates my enthusiasm for the field and my dedication to staying at the forefront of digital marketing trends.\\
Throughout my career, I have embraced opportunities to lead teams, develop innovative marketing solutions, and foster collaborative environments. My medium socioeconomic background has instilled in me a strong work ethic and a drive to excel, qualities that have been instrumental in my professional journey.\\
I am eager to bring my expertise in paid media and my passion for digital marketing to your team, contributing to innovative campaigns that drive success and growth. With a proven track record of delivering results and a relentless pursuit of excellence, I am excited about the opportunity to make a meaningful impact in your organization.
\par\endgroup
\end{prompt}

\twocolumn
\section{Implementation Details}
\label{sec:implementations}

\subsection{Explainability Methods}
\label{sub:explainability}

\paragraph{Concept classifiers.}
For all three datasets, we train a dedicated concept classifier that maps each (input, concept) pair to a discrete concept level and use it as a building block for all explanation methods. To ensure a fair comparison, all classifiers are trained on the same subset of 500 examples allocated to the explanation methods (using a 90\%--10\% train–validation split). Across datasets, we fine-tune the \texttt{microsoft/DeBERTa-v3-base} encoder from the Hugging Face \texttt{transformers} library.\footnote{\url{https://huggingface.co/microsoft/DeBERTa-v3-base}, \url{https://huggingface.co/docs/transformers}}
Each record–concept pair is converted into a templated input of the form \emph{``Concept: <concept>. Description: <text>''}, and the model predicts one of the concept’s discretized levels (2–4 values).

For the Violence dataset, we fine-tune for 4 epochs with a learning rate of $4\times10^{-5}$, a batch size of 4, a weight decay of 0.01, and 500 warmup steps, achieving 96.\% accuracy on the held-out test set. For the Disease dataset, we train for 3 epochs with a learning rate of $5\times10^{-5}$, a batch size of 8, a weight decay of 0.02, and 500 warmup steps, achieving 90.1\% accuracy. For the CV dataset, we fine-tune for 4 epochs with a learning rate of $3\times10^{-5}$, a batch size of 8, a weight decay of 0.01, and 500 warmup steps, achieving 94.4\% accuracy.

\paragraph{LEACE.}
We implement LEACE (Linear Erasure for Causal Effect) using the official \texttt{concept-erasure} library\footnote{\url{https://github.com/EleutherAI/concept-erasure}}, which provides the \texttt{LeaceFitter} object for estimating linear erasure operators. For each concept, we compute a separate LEACE erasure operator by iterating over the training split and extracting the model's final-layer hidden states. Concept labels are encoded using one-hot vectors, and each \texttt{LeaceFitter} is updated accordingly.
At inference time, we apply the learned erasure operator by registering a forward hook on the model's embedding layer, replacing the original embedding with its erased version for the target concept. Our implementation supports three backbone models: DeBERTa-v3-base, T5-base, and Qwen2.5-1.5B-Instruct,each loaded via the Hugging Face \texttt{transformers} and \texttt{peft} libraries.

\paragraph{ConceptShap}
For ConceptShap, we follow the protocol outlined by \cite{cebab2022} to ensure concept definitions remain consistent across all methods. First, we learn a vector representation for each concept using TCAV \cite{TCAV_Kim2018} implementation \footnote{\url{https://github.com/agil27/TCAV_PyTorch/tree/master}}. We then adapt a PyTorch implementation to ConceptShap to utilize these fixed concept vectors\footnote{\url{https://github.com/arnav-gudibande/conceptSHAP}}. Consistent with our LEACE setup, we support the same three backbone models loaded via Hugging Face.
\subsection{Explained Models}
\label{sub:explained}

The explanation methods operate on predictions generated by five models: 
DeBERTa-v3-base,\footnote{\url{https://huggingface.co/microsoft/DeBERTa-v3-base}} 
T5-base,\footnote{\url{https://huggingface.co/t5-base}} 
Qwen2.5-1.5B-Instruct,\footnote{\url{https://huggingface.co/Qwen/Qwen2.5-1.5B-Instruct}} 
GPT-4o,\footnote{\url{https://platform.openai.com/docs/models\#gpt-4o}} 
and LLaMA-3.1-Instruct.\footnote{\url{https://huggingface.co/meta-llama/Llama-3.1-8B-Instruct}} 
Each model is trained or prompted using a task-specific configuration.  
For reproducibility, Table~\ref{tab:explained-models-clean} reports the complete hyperparameter settings, implementation details, and predictive performance (accuracy and F1) for all trained models across the three datasets.

\begin{table*}[t]
\centering
\footnotesize
\begin{adjustbox}{width=0.95\textwidth}
\begin{tabular}{l|ccc|cc|l}
\toprule
\multicolumn{7}{c}{\cellcolor{Gray!40}\textbf{Workplace Violence Prediction – Explained Models}} \\
\midrule
\cellcolor{Gray!15}$\rightarrow$ \textbf{Model} &
\cellcolor{Gray!15}\textbf{LR} &
\cellcolor{Gray!15}\textbf{Batch} &
\cellcolor{Gray!15}\textbf{Epochs} &
\cellcolor{Gray!15}\textbf{Acc} &
\cellcolor{Gray!15}\textbf{F1} &
\cellcolor{Gray!15}\textbf{Notes} \\
\midrule
\textbf{DeBERTa-v3-base} &
$3\times10^{-5}$ & 8 & 5 &
73.75\% & 70.47\% &
Warmup 500, WD 0.01, linear scheduler \\[0.3em]

\textbf{T5-base} &
$5\times10^{-5}$ & 16 & 11 &
64.78\% & 57.47\% &
Weight decay 0.01, ``classify:'' prefix \\[0.3em]

\textbf{Qwen2.5-1.5B-Instruct} &
$5\times10^{-5}$ & 1 & 8 &
73.42\% & 71.20\% &
LoRA (r=16, alpha=32), GradAcc=8, WD=0.1 \\

\midrule
\midrule
\multicolumn{7}{c}{\cellcolor{Gray!40}\textbf{Disease Detection – Explained Models}} \\
\midrule
\cellcolor{Gray!15}$\rightarrow$ \textbf{Model} &
\cellcolor{Gray!15}\textbf{LR} &
\cellcolor{Gray!15}\textbf{Batch} &
\cellcolor{Gray!15}\textbf{Epochs} &
\cellcolor{Gray!15}\textbf{Acc} &
\cellcolor{Gray!15}\textbf{F1} &
\cellcolor{Gray!15}\textbf{Notes} \\
\midrule
\textbf{DeBERTa-v3-base} &
$3\times10^{-5}$ & 8 & 5 &
71.71\% & 71.69\% &
Warmup 500, WD 0.01, linear scheduler \\[0.3em]

\textbf{T5-base} &
$3\times10^{-4}$ & 16 & 10 &
70.39\% & 70.47\% &
Weight decay 0.01, ``classify:'' prefix \\[0.3em]

\textbf{Qwen2.5-1.5B-Instruct} &
$1\times10^{-4}$ & 1 & 8 &
62.83\% & 62.06\% &
LoRA (r=16, alpha=32), GradAcc=8, WD=0.1 \\

\midrule
\midrule
\multicolumn{7}{c}{\cellcolor{Gray!40}\textbf{CV Screening – Explained Models}} \\
\midrule
\cellcolor{Gray!15}$\rightarrow$ \textbf{Model} &
\cellcolor{Gray!15}\textbf{LR} &
\cellcolor{Gray!15}\textbf{Batch} &
\cellcolor{Gray!15}\textbf{Epochs} &
\cellcolor{Gray!15}\textbf{Acc} &
\cellcolor{Gray!15}\textbf{F1} &
\cellcolor{Gray!15}\textbf{Notes} \\
\midrule
\textbf{DeBERTa-v3-base} &
$5\times10^{-5}$ & 8 & 5 &
66.0\% & 65.05\% &
Warmup 500, WD 0.01, linear scheduler \\[0.3em]

\textbf{T5-base} &
$5\times10^{-5}$ & 16 & 9 &
70.0\% & 69.5\% &
Weight decay 0.01, ``classify:'' prefix \\[0.3em]

\textbf{Qwen2.5-1.5B-Instruct} &
$5\times10^{-5}$ & 1 & 8 &
49.33\% & 51.03\% &
LoRA (r=16, alpha=32), GradAcc=8, WD=0.1 \\

\bottomrule
\end{tabular}
\end{adjustbox}
\caption{Hyperparameters, implementation details, and predictive performance across all three datasets.}
\label{tab:explained-models-clean}
\end{table*}

\subsection{Prompts}
\label{sub:prompts}

\subsubsection{Explained Model Prompts}
\label{sub:prompts-classification-models}
To evaluate the explanation methods, we treat the five predictive models (DeBERTa, T5, Qwen2.5, GPT 4o, and LLaMA 3) as the models to be explained. Since these models differ in their interfaces and prompting requirements, we construct a dataset-specific input prompt for each one. 
Some models, such as DeBERTa, operate directly on the raw text, while instruction tuned models rely on natural language prompts that specify the task and the expected output format.

The full prompt templates appear in Table~\ref{tab:cv-model-prompts} for the CV dataset, Table~\ref{tab:violence-model-prompts} for the Violence dataset, and Table~\ref{tab:disease-model-prompts} for the Disease dataset.

\begin{table*}[h!]
\renewcommand{\arraystretch}{1.25}
\centering
\scriptsize
\caption{Prompt templates used for CV Explained models}
\label{tab:cv-model-prompts}
\begin{tabular}{p{0.15\textwidth}p{0.8\textwidth}}
\toprule
\cellcolor{Gray!15}\textbf{Model} & \cellcolor{Gray!15}\textbf{Input Format} \\
\midrule

\textbf{FT DeBERTa-v3-base} &
\begin{prompt}[label={box:cv-deberta-prompt}]{OliveGreen}{DeBERTa CV Classifier}
\footnotesize
\textit{No natural-language prompt is used. Input: \texttt{\{CV\_statement\}}}
\end{prompt}
\\[0.8em]

\textbf{FT T5-base} &
\begin{prompt}[label={box:cv-t5-prompt}]{OliveGreen}{T5 CV Classifier}
\footnotesize
\textit{classify: Rate the employee as 0 (Regular), 1 (Good), or 2 (Exceptional): \texttt{\{CV\_statement\}}}
\end{prompt}
\\[0.8em]

\textbf{FT Qwen2.5-1.5B-Instruct} &
\begin{prompt}[label={box:cv-qwen-prompt}]{OliveGreen}{Qwen CV Classifier}
\footnotesize
\textit{Classify CVs as 0-Regular, 1-Good, or 2-Exceptional based on professional (e.g.,
experience, education, achievements, volunteering) and demographic information
(e.g., gender, age, race, socioeconomic status). \texttt{\{CV\_statement\}}}
\end{prompt}
\\[0.8em]

\textbf{Zero-shot} &
\begin{prompt}[label={box:cv-gpt4o-prompt}]{OliveGreen}{GPT-4o CV Classifier}
\footnotesize
\textit{\textbf{System:} You are an HR specialist tasked with screening CVs by evaluating job candidates based on their self-statement. In the self-statement, candidates typically provide
both professional details (e.g., experience, education, achievements, volunteering)
and demographic information (e.g., gender, age, race, socioeconomic status). Use
both types of information, along with your world knowledge and understanding of
what makes a successful employee, to make a well-informed evaluation. We have a
large pool of candidates, all of whom are already considered a good fit for the role.
Your task is to carefully evaluate each candidate based on their self-statement and
assign one of the following scores:
0: A solid and competent candidate who meets the role's requirements.
1: A promising candidate with potential, demonstrating notable qualities or
attributes that suggest they could become exceptional with further development.
2: An outstanding candidate, one of a kind, with extraordinary achievements and
qualities that make them an ideal hire. Use your understanding of workplace
success and the information provided in the self-statement to make your decision.
Return only a single character: 0, 1, or 2.\\
\textbf{User:} The job role is: \{Persona\_job\}. The CV self-statement is: \{CV\_statement\}.}
\end{prompt}
\\
\bottomrule
\end{tabular}
\end{table*}

\begin{table*}[h!]
\renewcommand{\arraystretch}{1.25}
\centering
\scriptsize
\caption{Prompt templates used for Violence Explained models}
\label{tab:violence-model-prompts}
\begin{tabular}{p{0.15\textwidth}p{0.8\textwidth}}
\toprule
\cellcolor{Gray!15}\textbf{Model} & \cellcolor{Gray!15}\textbf{Input Format} \\
\midrule

\textbf{FT DeBERTa-v3-base} &
\begin{prompt}[label={box:violence-deberta-prompt}]{OliveGreen}{DeBERTa Violence Classifier}
\footnotesize
\textit{No natural-language prompt is used. Input: \texttt{\{Dialogue\}}}
\end{prompt}
\\[0.8em]

\textbf{FT T5-base} &
\begin{prompt}[label={box:violence-t5-prompt}]{OliveGreen}{T5 Violence Classifier}
\footnotesize
\textit{classify: Receive a dialogue between HR and an employee. Infer if the employee experienced workplace violence. 0=No, 1=Verbal, 2=Physical. The dialogue: \texttt{\{Dialogue\}}}
\end{prompt}
\\[0.8em]

\textbf{FT Qwen2.5-1.5B-Instruct} &
\begin{prompt}[label={box:violence-qwen-prompt}]{OliveGreen}{Qwen Violence Classifier}
\footnotesize
\textit{Given a dialogue between an HR representative and a nurse, infer whether the nurse experienced or will experience workplace violence based on Gender, Age, Race, Years as a Nurse, License Type, Department, and Workforce Position. Classify as: 0=No violence, 1=Verbal violence, 2=Physical violence. Dialogue: \texttt{\{Dialogue\}}}
\end{prompt}
\\[0.8em]

\textbf{Zero-shot} &
\begin{prompt}[label={box:violence-gpt4o-prompt}]{OliveGreen}{GPT-4o Violence Classifier}
\footnotesize
\textit{\textbf{System:} 
You are a specialist responsible for assessing workplace violence risks in nursing environments.
Analyze a dialogue between an HR representative and a nurse to identify key contextual factors about the nurse and their workplace environment. These factors include Gender, Age, Race, Years of Experience, License Type, Department, and Position in the Workforce Hierarchy, which are known to be linked to the likelihood of experiencing workplace violence. Based on these aspects, determine the appropriate risk level:
0: Standard workplace risk – Similar to the general nursing workforce, with no strong indicators of increased risk.
1: Elevated risk – A higher likelihood of verbal abuse, threats, harassment, or intimidation. 
2: High risk – A significant likelihood of physical harm or the threat of physical harm.
Return only a single character: 0, 1, or 2.\\
\textbf{User:} Dialogue: \{Dialogue\}.}
\end{prompt}
\\
\bottomrule
\end{tabular}
\end{table*}

\begin{table*}[h!]
\renewcommand{\arraystretch}{1.25}
\centering
\scriptsize
\caption{Prompt templates used for Disease Explained models}
\label{tab:disease-model-prompts}
\begin{tabular}{p{0.15\textwidth}p{0.8\textwidth}}
\toprule
\cellcolor{Gray!15}\textbf{Model} & \cellcolor{Gray!15}\textbf{Input Format} \\
\midrule

\textbf{FT DeBERTa-v3-base} &
\begin{prompt}[label={box:disease-deberta-prompt}]{OliveGreen}{DeBERTa Disease Classifier}
\footnotesize
\textit{No natural-language prompt is used. Input: \texttt{\{Patient\_consultation\}}}
\end{prompt}
\\[0.8em]

\textbf{FT T5-base} &
\begin{prompt}[label={box:disease-t5-prompt}]{OliveGreen}{T5 Disease Classifier}
\footnotesize
\textit{classify: Diagnose the patient based on their symptoms (symptoms such as dizziness, sensitivity to light, headache, nasal congestion, facial pain or pressure, fever, and general weakness. Your goal is to classify the most probable diagnosis based on these symptoms. The possible classifications are:
0: Migraine – Typically includes dizziness, sensitivity to light, and headache. 
1: Sinusitis – Commonly presents with nasal congestion, facial pain or pressure, fever and headache. 
2: Influenza – Characterized by fever, general weakness, nasal congestion and headache.
Patient's complaint: \texttt{\{Patient\_consultation\}} 
Return only a single character: 0, 1, or 2.}
\end{prompt}
\\[0.8em]

\textbf{FT Qwen2.5-1.5B-Instruct} &
\begin{prompt}[label={box:disease-qwen-prompt}]{OliveGreen}{Qwen Disease Classifier}
\footnotesize
\textit{
You are a medical specialist diagnosing patients based on their reported symptoms. Each complaint describes symptoms such as dizziness, sensitivity to light, headache, nasal congestion, facial pain or pressure, fever, and general weakness. Analyze the complaint and classify the most probable diagnosis: 0: Migraine, 1: Sinusitis, 2: Influenza. Return only a single character: 0, 1, or 2. \texttt{\{Patient\_consultation\}}}
\end{prompt}
\\[0.8em]

\textbf{Zero-shot} &
\begin{prompt}[label={box:disease-gpt4o-prompt}]{OliveGreen}{GPT-4o Disease Classifier}
\footnotesize
\textit{\textbf{System:} You are a medical specialist responsible for diagnosing patients based on their reported symptoms. Each patient provides a complaint describing their condition, which includes symptoms such as dizziness, sensitivity to light, headache, nasal congestion, facial pain or pressure, fever, and general weakness. Your task is to carefully analyze the complaint and determine the most probable diagnosis from the following categories: 0: Migraine, 1: Sinusiti, 2: Influenza. Use your medical knowledge to assess the connection between the symptoms described and the most likely underlying disease. Return only a single character: 0, 1, or 2.\\
\textbf{User:} Patient's complaint: \{Patient\_consultation\}.}
\end{prompt}
\\
\bottomrule
\end{tabular}
\end{table*}

\subsubsection{CF Generation method}
\label{sub:prompts-CF-Generation-method}

In counterfactual generation, we evaluated four prompt formulations that operationalize distinct causal assumptions for generating counterfactuals. Each prompt reflects a different constraint on which concepts may or may not change to maintain causal coherence~\ref{tab:cf-generation-prompts}.\\[0.5em]

\begin{table*}[h!]
\renewcommand{\arraystretch}{1.25}
\centering
\scriptsize
\caption{Prompt formulations used for counterfactual generation test}
\label{tab:cf-generation-prompts}
\begin{tabular}{p{0.1\textwidth}p{0.89\textwidth}}
\toprule
\cellcolor{Gray!15}\textbf{Prompt Type} & \cellcolor{Gray!15}\textbf{Full Prompt Text} \\
\midrule

\textbf{Only Change} &
\begin{prompt}[label={box:v1-exact-matching-prompt}]{OliveGreen}{Only Change Prompt}
\footnotesize
\textit{
\textbf{Prompt Instruction:}\\
I'm providing a CV statement from the LIBERTy dataset. Update it by modifying only the `\{concept\}` concept.\\
---- Input CV Statement ----\\
\{text\}\\
---- Instruction ----\\
The candidate's `\{concept\}` is `\{old\_value\_text\}`. Change it to `\{new\_value\_text\}` while keeping all other aspects unchanged.\\
---- Edited CV Statement ----\\
Return only the updated CV statement with no additional text.
}
\end{prompt} \\[0.8em]

\textbf{Fix Confounders (Confounders Focus)} &
\begin{prompt}[label={box:v3-confounders-prompt}]{OliveGreen}{Fix confounder Prompt}
\footnotesize
\textit{
\textbf{Prompt Instruction:}\\
I'm providing a CV statement from the LIBERTy dataset. Your task is to update it by modifying the `\{concept\}` concept.\\
---- Input CV Statement ----\\
\{text\}\\
---- Instruction ----\\
The candidate's `\{concept\}` is `\{old\_value\_text\}`. Change it to `\{new\_value\_text\}`.\\
The following concepts are \textbf{confounders} and must \textbf{not} be changed: \{', '.join(confounders)\}.\\
---- Edited CV Statement ----\\
Return only the updated CV statement with no additional text.
}
\end{prompt} \\[0.8em]

\textbf{Fix All (Flexible Change)} &
\begin{prompt}[label={box:v4-flexible-prompt}]{OliveGreen}{Fix all Prompt}
\footnotesize
\textit{
\textbf{Prompt Instruction:}\\
I'm providing a CV statement from the LIBERTy dataset. Your task is to update it by modifying the `\{concept\}` concept.\\
---- Input CV Statement ----\\
\{text\}\\
---- Instruction ----\\
The candidate's `\{concept\}` is `\{old\_value\_text\}`. Change it to `\{new\_value\_text\}`.\\
The CV statement includes the following concepts: \{', '.join(all\_concepts)\}.\\
Some of these concepts are causally linked to `\{concept\}` and may require adjustments to maintain logical consistency.\\
---- Edited CV Statement ----\\
Return only the updated CV statement with no additional text.
}
\end{prompt} \\[0.8em]

\textbf{Mediators and Confounders (Causal Framework)} &
\begin{prompt}[label={box:v6-causal-framework-prompt}]{OliveGreen}{Mediators and Confounders Prompt}
\footnotesize
\textit{
\textbf{Prompt Instruction:}\\
I'm providing a CV statement from the LIBERTy dataset. Your task is to update it by modifying the `\{concept\}` concept.\\
---- Input CV Statement ----\\
\{text\}\\
---- Instruction ----\\
The candidate's `\{concept\}` is `\{old\_value\_text\}`. Change it to `\{new\_value\_text\}`.\\
The following concepts are \textbf{confounders}, meaning they must remain unchanged: \{', '.join(confounders)\}.\\
The following concepts are \textbf{mediators}, meaning they are causally influenced by `\{concept\}` and may require adjustments to maintain logical consistency: \{', '.join(mediators)\}.\\
---- Edited CV Statement ----\\
Return only the updated CV statement with no additional text.
}
\end{prompt} \\
\bottomrule
\end{tabular}
\end{table*}

\section{Additional Results}
\label{sec:additional}

In this section, we present the complete results for the three core experiments conducted in this work: 
\emph{Local Explainability}, \emph{Global Explainability}, and \emph{Concept Sensitivity Analysis}.  
Each experiment is evaluated across all three datasets, \emph{Workplace Violence Prediction}, \emph{Disease Detection}, and \emph{CV Screening}, and the tables below provide the full quantitative outcomes that complement the summaries reported in the main text.  
Specifically: 

\begin{enumerate}
    \item \textbf{Local Explainability (Table~\ref{tab:complete_local})}:  
    This table reports the full ICaCE Error-Distance (ED) and Order-Faithfulness (OF) scores for all explanation methods and all models, across each dataset.  

    \item \textbf{Global Explainability (Table~\ref{tab:complete_global})}: 
    This table presents the complete set of global OF scores, aggregated across all examples.  

    \item \textbf{Concept Sensitivity Analysis (Table~\ref{tab:complete_sensitivity})}:  
    This table reports full sensitivity scores for all concepts, models, and datasets.  
    It includes ICaCE-based sensitivity magnitudes and ground-truth sensitivities derived from structural causal models.
\end{enumerate}

\begin{table*}[t]
\centering
\footnotesize
\begin{adjustbox}{width=0.8\textwidth}
\begin{tabular}{l|cc|cc|cc|cc|cc|cc}
\toprule
\multicolumn{13}{c}{\cellcolor{Gray!40}\textbf{Workplace Violence Prediction}} \\
\midrule
 \cellcolor{Gray!15} $\rightarrow$ \textbf{Model}  &
\multicolumn{2}{c|}{\cellcolor{Gray!15}\textbf{Average}} &
\multicolumn{2}{c|}{\cellcolor{Gray!15}\textbf{DeBERTa-v3 }} &
\multicolumn{2}{c|}{\cellcolor{Gray!15}\textbf{T5
}} &
\multicolumn{2}{c|}{\cellcolor{Gray!15}\textbf{Qwen-2.5}} &
\multicolumn{2}{c|}{\cellcolor{Gray!15}\textbf{Llama-3.1}} &
\multicolumn{2}{c}{\cellcolor{Gray!15}\textbf{GPT-4o}} \\
\cellcolor{Gray!15} $\downarrow$ \textbf{Method} & \cellcolor{Gray!15} ED & \cellcolor{Gray!15} OF & \cellcolor{Gray!15} ED & \cellcolor{Gray!15} OF & \cellcolor{Gray!15} ED & \cellcolor{Gray!15} OF & \cellcolor{Gray!15} ED & \cellcolor{Gray!15} OF & \cellcolor{Gray!15} ED & \cellcolor{Gray!15} OF & \cellcolor{Gray!15} ED & \cellcolor{Gray!15} OF \\
\midrule
\textit{CF Gen} & 0.47 & 0.58 & 0.39 & 0.71 & 0.37 & 0.67 & 0.54 & 0.64 & 0.51 & 0.32 & 0.52 & 0.57 \\
\textit{Approx} & 0.41 & 0.71 & 0.34 & 0.80 & 0.31 & 0.76 & 0.48 & 0.76 & 0.42 & 0.51 & 0.51 & 0.70 \\
\textit{ConVecs} & 0.40 & 0.73 & 0.27 & 0.86 & 0.34 & 0.77 & 0.44 & \textbf{0.79} & 0.42 & 0.51 & 0.51 & 0.71 \\
\textit{ST Match} & 0.51 & 0.63 & 0.53 & 0.68 & 0.44 & 0.66 & 0.65 & 0.65 & 0.41 & 0.50 & 0.52 & 0.68 \\
\textit{PT Match} & 0.51 & 0.64 & 0.53 & 0.67 & 0.44 & 0.66 & 0.65 & 0.65 & \textbf{0.37} & \textbf{0.56} & 0.57 & 0.64 \\
\textit{FT Match} & \textbf{0.32} & \textbf{0.84} & \textbf{0.11} & \textbf{0.93} & \textbf{0.23} & \textbf{0.83} & \textbf{0.39} & \textbf{0.79} & 0.38 & 0.52 & \textbf{0.46} & \textbf{0.72} \\
\midrule
\midrule
\multicolumn{13}{c}{\cellcolor{Gray!40}\textbf{Disease Detection}} \\
\midrule
\cellcolor{Gray!15}  $\rightarrow$ \textbf{Model}  &
\multicolumn{2}{c|}{\cellcolor{Gray!15}\textbf{Average}} &
\multicolumn{2}{c|}{\cellcolor{Gray!15}\textbf{DeBERTa-v3 }} &
\multicolumn{2}{c|}{\cellcolor{Gray!15}\textbf{T5
}} &
\multicolumn{2}{c|}{\cellcolor{Gray!15}\textbf{Qwen-2.5}} &
\multicolumn{2}{c|}{\cellcolor{Gray!15}\textbf{Llama-3.1}} &
\multicolumn{2}{c}{\cellcolor{Gray!15}\textbf{GPT-4o}} \\
\cellcolor{Gray!15} $\downarrow$ \textbf{Method} & \cellcolor{Gray!15} ED & \cellcolor{Gray!15} OF & \cellcolor{Gray!15} ED & \cellcolor{Gray!15} OF & \cellcolor{Gray!15} ED & \cellcolor{Gray!15} OF & \cellcolor{Gray!15} ED & \cellcolor{Gray!15} OF & \cellcolor{Gray!15} ED & \cellcolor{Gray!15} OF & \cellcolor{Gray!15} ED & \cellcolor{Gray!15} OF \\
\midrule
\textit{CF Gen} & 0.67 & 0.36 & 0.63 & 0.47 & 0.54 & 0.48 & 0.59 & 0.46 & 0.78 & 0.10 & 0.79 & 0.31 \\
\textit{Approx} & 0.48 & 0.69 & 0.43 & 0.74 & 0.43 & 0.71 & 0.51 & 0.66 & 0.53 & 0.63 & 0.50 & 0.69 \\
\textit{ConVecs} & 0.44 & 0.70 & 0.38 & 0.74 & 0.41 & 0.72 & 0.46 & 0.67 & 0.50 & 0.63 & 0.47 & 0.72 \\
\textit{ST Match} & 0.46 & 0.69 & 0.43 & 0.72 & 0.41 & 0.71 & 0.45 & 0.68 & \textbf{0.44} & 0.65 & 0.56 & 0.70 \\
\textit{PT Match} & 0.52 & 0.65 & 0.49 & 0.70 & 0.49 & 0.66 & 0.49 & 0.65 & 0.49 & 0.60 & 0.65 & 0.66 \\
\textit{FT Match} & \textbf{0.36} & \textbf{0.75} & \textbf{0.18} & \textbf{0.86} & \textbf{0.31} & \textbf{0.78} & \textbf{0.39} & \textbf{0.73} & \textbf{0.44} & \textbf{0.66} & \textbf{0.46} & \textbf{0.73} \\
\textit{LEACE} & 0.65 & 0.46 & 0.62 & 0.42 & 0.46 & 0.55 & 0.87 & 0.41 & -- & -- & -- & -- \\
\midrule
\midrule
\multicolumn{13}{c}{\cellcolor{Gray!40}\textbf{CV Screening}} \\
\midrule
 \cellcolor{Gray!15} $\rightarrow$ \textbf{Model}  &
\multicolumn{2}{c|}{\cellcolor{Gray!15}\textbf{Average}} &
\multicolumn{2}{c|}{\cellcolor{Gray!15}\textbf{DeBERTa-v3 }} &
\multicolumn{2}{c|}{\cellcolor{Gray!15}\textbf{T5
}} &
\multicolumn{2}{c|}{\cellcolor{Gray!15}\textbf{Qwen-2.5}} &
\multicolumn{2}{c|}{\cellcolor{Gray!15}\textbf{Llama-3.1}} &
\multicolumn{2}{c}{\cellcolor{Gray!15}\textbf{GPT-4o}} \\
\cellcolor{Gray!15} $\downarrow$ \textbf{Method} & \cellcolor{Gray!15} ED & \cellcolor{Gray!15} OF & \cellcolor{Gray!15} ED & \cellcolor{Gray!15} OF & \cellcolor{Gray!15} ED & \cellcolor{Gray!15} OF & \cellcolor{Gray!15} ED & \cellcolor{Gray!15} OF & \cellcolor{Gray!15} ED & \cellcolor{Gray!15} OF & \cellcolor{Gray!15} ED & \cellcolor{Gray!15} OF \\
\midrule
\textit{CF Gen}   & 0.52 & 0.52 & 0.48 & 0.58 & 0.49 & 0.55 & 0.73 & 0.48 & 0.47 & 0.39 & \textbf{0.43} & 0.60 \\
\textit{Approx}   & 0.46 & 0.67 & 0.36 & 0.74 & 0.33 & 0.71 & 0.51 & 0.69 & 0.50 & 0.56 & 0.58 & 0.63 \\
\textit{ConVecs}  & 0.47 & 0.66 & 0.38 & 0.75 & 0.39 & 0.71 & 0.50 & 0.67 & 0.52 & 0.53 & 0.57 & 0.62 \\
\textit{ST Match} & 0.50 & 0.62 & 0.52 & 0.67 & 0.48 & 0.63 & 0.56 & 0.64 & 0.41 & 0.56 & 0.52 & 0.62 \\
\textit{PT Match} & 0.50 & 0.63 & 0.53 & 0.68 & 0.49 & 0.63 & 0.54 & 0.65 & \textbf{0.40} & 0.56 & 0.55 & 0.63 \\
\textit{FT Match} & \textbf{0.35} & \textbf{0.72} & \textbf{0.19} & \textbf{0.86} & \textbf{0.26} & \textbf{0.78} & \textbf{0.40} & \textbf{0.73} & 0.42 & \textbf{0.57} & 0.50 & \textbf{0.65} \\

\bottomrule
\end{tabular}
\end{adjustbox}
\caption{\textbf{Local Explainability -- Full Results:} We report the Average ICaCE Error-Distance ($\overline{\mathrm{ED}}$, $\downarrow$ is better) and Average ICaCE Order-Faithfulness ($\overline{\mathrm{OF}}$, $\uparrow$ is better). The \textbf{Average} column reports the mean across five explained models and three datasets.} 
\label{tab:complete_local}
\end{table*}

\begin{table*}[t]
\centering
\footnotesize
\begin{adjustbox}{width=0.8\textwidth}
\begin{tabular}{l|c|ccccc}
\toprule
\multicolumn{7}{c}{\cellcolor{Gray!40}\textbf{Workplace Violence Prediction}} \\
\midrule
\cellcolor{Gray!15} $\rightarrow$ \textbf{Model}  &
\cellcolor{Gray!15}\textbf{Average} &
\cellcolor{Gray!15}\textbf{DeBERTa-v3 } &
\cellcolor{Gray!15}\textbf{T5} &
\cellcolor{Gray!15}\textbf{Qwen-2.5} &
\cellcolor{Gray!15}\textbf{Llama-3.1} &
\cellcolor{Gray!15}\textbf{GPT-4o}  \\
\midrule
\textit{CF Gen} & 0.772 & 0.810 & 0.857 & 0.762 & 0.476 & \textbf{0.952} \\
\textit{Approx} & 0.781 & 0.810 & 0.905 & 0.810 & 0.524 & 0.810 \\
\textit{ConVecs} & 0.829 & 0.905 & 0.905 & \textbf{0.952} & 0.524 & 0.857 \\
\textit{ST Match} & 0.743 & 0.714 & 0.810 & 0.762 & 0.571 & 0.857 \\
\textit{PT Match} & 0.762 & 0.714 & 0.810 & 0.857 & 0.714 & 0.714 \\
\textit{FT Match} & \textbf{0.857} & \textbf{1.000} & \textbf{1.000} & 0.762 & 0.571 & \textbf{0.952} \\
\textit{ConceptSHAP} & 0.444 & 0.381 & 0.333 & 0.619 & -- & -- \\
\midrule
\midrule
\multicolumn{7}{c}{\cellcolor{Gray!40}\textbf{Disease Detection}} \\
\midrule
\cellcolor{Gray!15} $\rightarrow$ \textbf{Model}  &
\cellcolor{Gray!15}\textbf{Average} &
\cellcolor{Gray!15}\textbf{DeBERTa-v3 } &
\cellcolor{Gray!15}\textbf{T5} &
\cellcolor{Gray!15}\textbf{Qwen-2.5} &
\cellcolor{Gray!15}\textbf{Llama-3.1} &
\cellcolor{Gray!15}\textbf{GPT-4o}  \\
\midrule
\textit{CF Gen} & 0.476 & 0.619 & 0.333 & 0.333 & 0.524 & 0.571 \\
\textit{Approx} & 0.838 & \textbf{1.000} & \textbf{0.810} & 0.714 & 0.810 & 0.857 \\
\textit{ConVecs} & 0.876 & 0.905 & \textbf{0.810} & \textbf{0.905} & 0.857 & 0.905 \\
\textit{ST Match} & 0.790 & 0.762 & 0.714 & 0.619 & \textbf{0.952} & 0.905 \\
\textit{PT Match} & 0.629 & 0.762 & 0.381 & 0.524 & 0.810 & 0.667 \\
\textit{FT Match} & \textbf{0.877} & 0.905 & \textbf{0.810} & 0.857 & 0.905 & \textbf{0.952} \\
\textit{LEACE} & 0.619 & 0.667 & 0.571 & 0.619 & -- & -- \\
\textit{ConceptSHAP} & 0.333 & 0.524 & 0.190 & 0.286 & -- & -- \\
\midrule
\midrule
\multicolumn{7}{c}{\cellcolor{Gray!40}\textbf{CV Screening}} \\
\midrule
\cellcolor{Gray!15} $\rightarrow$ \textbf{Model}  &
\cellcolor{Gray!15}\textbf{Average} &
\cellcolor{Gray!15}\textbf{DeBERTa-v3 } &
\cellcolor{Gray!15}\textbf{T5} &
\cellcolor{Gray!15}\textbf{Qwen-2.5} &
\cellcolor{Gray!15}\textbf{Llama-3.1} &
\cellcolor{Gray!15}\textbf{GPT-4o} \\
\midrule
\textit{CF Gen} & 0.599 & 0.429 & 0.643 & 0.464 & 0.607 & 0.750 \\
\textit{Approx} & 0.685 & 0.643 & 0.750 & 0.714 & 0.571 & 0.750 \\
\textit{ConVecs} & 0.750 & 0.714 & 0.786 & 0.821 & 0.643 & \textbf{0.786} \\
\textit{ST Match} & 0.650 & 0.607 & 0.464 & 0.643 & 0.750 & \textbf{0.786} \\
\textit{PT Match} & 0.671 & 0.607 & 0.464 & 0.750 & 0.750 & \textbf{0.786} \\
\textit{FT Match} & \textbf{0.783} & \textbf{0.821} & \textbf{0.857} & \textbf{0.857} & \textbf{0.786} & 0.643 \\
\textit{ConceptSHAP} & 0.448 & 0.536 & 0.500 & 0.357 & -- & -- \\
\bottomrule
\end{tabular}
\end{adjustbox}
\caption{\textbf{Global Explainability – Full Results:} Mean Order-Faithfulness ($\overline{\mathrm{OF}}$, $\uparrow$ is better) for global explanations of each model and dataset. Bolded values mark the best-performing method per column.}
\label{tab:complete_global}
\end{table*}

\begin{table*}[t]
\centering
\footnotesize
\begin{adjustbox}{width=0.97\textwidth}
\begin{tabular}{lcccccccc}
\toprule

\multicolumn{9}{c}{\cellcolor{Gray!40}\textbf{Workplace Violence Prediction}} \\
\cellcolor{Gray!15}\textbf{Model} & 
\cellcolor{Gray!15}\textbf{Race} & 
\cellcolor{Gray!15}\textbf{Gender} & 
\cellcolor{Gray!15}\textbf{Age} & 
\cellcolor{Gray!15}\textbf{Seniority} & 
\cellcolor{Gray!15}\textbf{Department} & 
\cellcolor{Gray!15}\textbf{License} & 
\cellcolor{Gray!15}\textbf{Tenure} &
\cellcolor{Gray!15}\textbf{—} \\

DeBERT-v3 
 & 0.350 & 1.192 & 0.758 & 0.831 & 0.595 & \textbf{0.595} & 0.525 & \textemdash \\
T5
       & \textbf{0.421} & 0.743 & 0.512 & \textbf{0.645} & 0.569 & 0.452 & 0.307 & \textemdash \\
Qwen-2.5 & 0.691 & \textbf{1.314} & \textbf{1.045} & 0.713 & 1.308 & 0.656 & \textbf{0.597} & \textemdash \\
Llama-3.1    & 0.224 & 0.227 & 0.226 & 0.208 & 0.227 & 0.211 & 0.211 & \textemdash \\
GPT-4o & 0.724 & 0.594 & 0.300 & 0.413 & \textbf{1.203} & 0.279 & 0.256 & \textemdash \\
\rowcolor{Goldenrod!80}
\textbf{True Effect} & 
0.484 & 1.271 & 1.154 & 0.560 & 1.232 & 0.572 & 0.613 & \textemdash \\

\multicolumn{9}{c}{\cellcolor{Gray!40}\textbf{Disease Detection}} \\
\cellcolor{Gray!15}\textbf{Model} & 
\cellcolor{Gray!15}\textbf{Dizzy} & 
\cellcolor{Gray!15}\textbf{Facial Pain} & 
\cellcolor{Gray!15}\textbf{Fever} & 
\cellcolor{Gray!15}\textbf{Weakness} & 
\cellcolor{Gray!15}\textbf{Headache} & 
\cellcolor{Gray!15}\textbf{Nasal Congestion} & 
\cellcolor{Gray!15}\textbf{Light Sensitivity} &
\cellcolor{Gray!15}\textbf{—} \\

DeBERT-v3 
 & 0.505 & 0.593 & 0.243 & 0.415 & 0.398 & 0.395 & 0.693 & \textemdash \\
T5
       & 0.352 & 0.678 & 0.284 & 0.376 & 0.530 & 0.506 & 0.745 & \textemdash \\
Qwen-2.5     & 0.495 & 0.710 & 0.383 & 0.512 & 0.426 & 0.443 & 0.679 & \textemdash \\
Llama-3.1    & 0.487 & 0.442 & 0.474 & 0.332 & 0.364 & 0.587 & 0.519 & \textemdash \\
GPT-4o   & 0.364 & 0.644 & 0.504 & 0.215 & 0.369 & 0.684 & 0.879 & \textemdash \\

\multicolumn{9}{c}{\cellcolor{Gray!40}\textbf{CV Screening}} \\
\cellcolor{Gray!15}\textbf{Model} & 
\cellcolor{Gray!15}\textbf{Race} & 
\cellcolor{Gray!15}\textbf{Gender} & 
\cellcolor{Gray!15}\textbf{Age} & 
\cellcolor{Gray!15}\textbf{Education} & 
\cellcolor{Gray!15}\textbf{Socioeconomic} & 
\cellcolor{Gray!15}\textbf{Volunteering} & 
\cellcolor{Gray!15}\textbf{Experience} & 
\cellcolor{Gray!15}\textbf{Certificates} \\

DeBERT-v3 
& \textbf{0.715} & 0.432 & \textbf{0.613} & \textbf{1.297} & 0.245 & \textbf{0.391} & 0.285 & \textbf{0.732} \\
T5
       & 0.742 & 0.398 & 0.513 & 1.143 & 0.086 & 0.066 & 0.168 & 0.443 \\
Qwen-2.5 & 0.522 & \textbf{0.361} & 0.503 & 0.799 & 0.354 & 0.335 & \textbf{0.756} & 0.425 \\
Llama-3.1    & 0.374 & 0.283 & 0.397 & 0.437 & 0.336 & 0.349 & 0.381 & 0.329 \\
GPT-4o & 0.417 & 0.208 & 0.355 & 0.679 & \textbf{0.237} & 0.251 & 0.727 & 0.227 \\
\rowcolor{Goldenrod!80}
\textbf{True Effect} & 
0.636 & 0.369 & 0.913 & 1.357 & 0.209 & 0.586 & 0.866 & 0.599 \\

\bottomrule
\end{tabular}
\end{adjustbox}
\caption{\textbf{Concept Sensitivity Analysis -- Full Results:} We report concept sensitivity as follows: for each example and concept change, we compute ICaCE values and sum their absolute magnitudes across all output classes. The final concept score is then the average of this quantity across all examples and changes. We also report the ground-truth sensitivity of $Y$ from the SCMs, except in the Disease Detection dataset, where $Y$ (the disease) is the parent of the concepts (symptoms). } 
\label{tab:complete_sensitivity}
\end{table*}


\end{document}